\documentclass{article}

\usepackage[main, final]{neurips_2026}

\usepackage[utf8]{inputenc}
\usepackage[T1]{fontenc}
\usepackage[hypertexnames=false]{hyperref}
\hypersetup{pdftitle={Beyond Copy-Paste: How Well Do Subject-Driven Video Models Understand Their Subjects?},
  pdfauthor={Zun Wang, Kenan Deng, Daniel Blackburn, Linlin Lu, Junbang Liang, Yu Lou, Mohit Bansal, Shan Yang},
  colorlinks=false,pdfborder={0 0 0}}
\usepackage{url}
\usepackage{booktabs}
\usepackage{amsfonts}
\usepackage{amsmath,amssymb}
\usepackage{bm}
\usepackage{nicefrac}
\usepackage{microtype}
\usepackage{xcolor}
\usepackage[normalem]{ulem}
\usepackage{graphicx}
\usepackage{multirow}
\usepackage[capitalize,noabbrev]{cleveref}
\usepackage[breakable]{tcolorbox}
\usepackage{float}

\title{Beyond Copy-Paste: How Well Do Subject-Driven Video Models Understand Their Subjects?}

\author{\textbf{Zun Wang}\textsuperscript{1}\thanks{Work done during an internship at Amazon.}
  \quad \textbf{Kenan Deng}\textsuperscript{2}
  \quad \textbf{Daniel Blackburn}\textsuperscript{2}
  \quad \textbf{Linlin Lu}\textsuperscript{2}\\
  \textbf{Junbang Liang}\textsuperscript{2}
  \quad \textbf{Yu Lou}\textsuperscript{2}
  \quad \textbf{Mohit Bansal}\textsuperscript{1}
  \quad \textbf{Shan Yang}\textsuperscript{2}\\[0.5em]
  \textsuperscript{1}University of North Carolina at Chapel Hill
  \qquad
  \textsuperscript{2}Amazon
}

\begin{document}

\maketitle

\begin{abstract}
Subject-driven video generation aims to produce videos that faithfully incorporate user-provided reference subjects.
Current evaluation relies on free generation followed by reference-similarity scoring, which rewards a fundamental shortcut: models can obtain high scores by reproducing visible cues from the references while avoiding transformations that would test whether the subject is truly preserved beyond the conditioning input.
We design \emph{anchored evaluation}: controlled settings in which generation is anchored to held-out targets of the same subject rather than left unconstrained.
Each probe withholds target content from the model's conditioning, so success requires using the references to infer content not directly visible, rather than simply copy-pasting what was given.
We instantiate two probes.
(1) \emph{Viewpoint control} specifies target geometry via depth-based warping, leaving occluded regions that the model must complete using subject appearance from the references.
(2) \emph{Degradation--recovery} corrupts held-out videos showing the subject in real-world contexts that differ from the reference images, then recovers them conditioned on those references; recovery requires cross-context generalization of identity and inference of state dynamics.
Both probes require only lightweight modifications to the inference loop of flow-matching models, need no retraining, and generalize across architectures.
Evaluation across seven diverse, strong models reveals that rankings shift substantially: VACE-14B ranks first conventionally but Phantom-14B, ranked last, leads under anchored evaluation. Moreover, conventional scores strongly correlate with the improvement from providing ground-truth references (Pearson $r = 0.91$), suggesting they largely measure copying ability.
This interpretation is further supported by 2,000 human comparisons, which show that anchored scores align substantially better with human judgments than conventional scores.
Fine-grained analysis further reveals distinct failure modes: copy-paste models drift at unseen viewpoints, and no model adaptively balances copying and generalization.
Code and data will be publicly released at \url{https://github.com/wz0919/AnchorEvaluation}.
\end{abstract}

\section{Introduction}
\label{sec:intro}

\begin{figure*}[t]
\centering
\includegraphics[width=\textwidth]{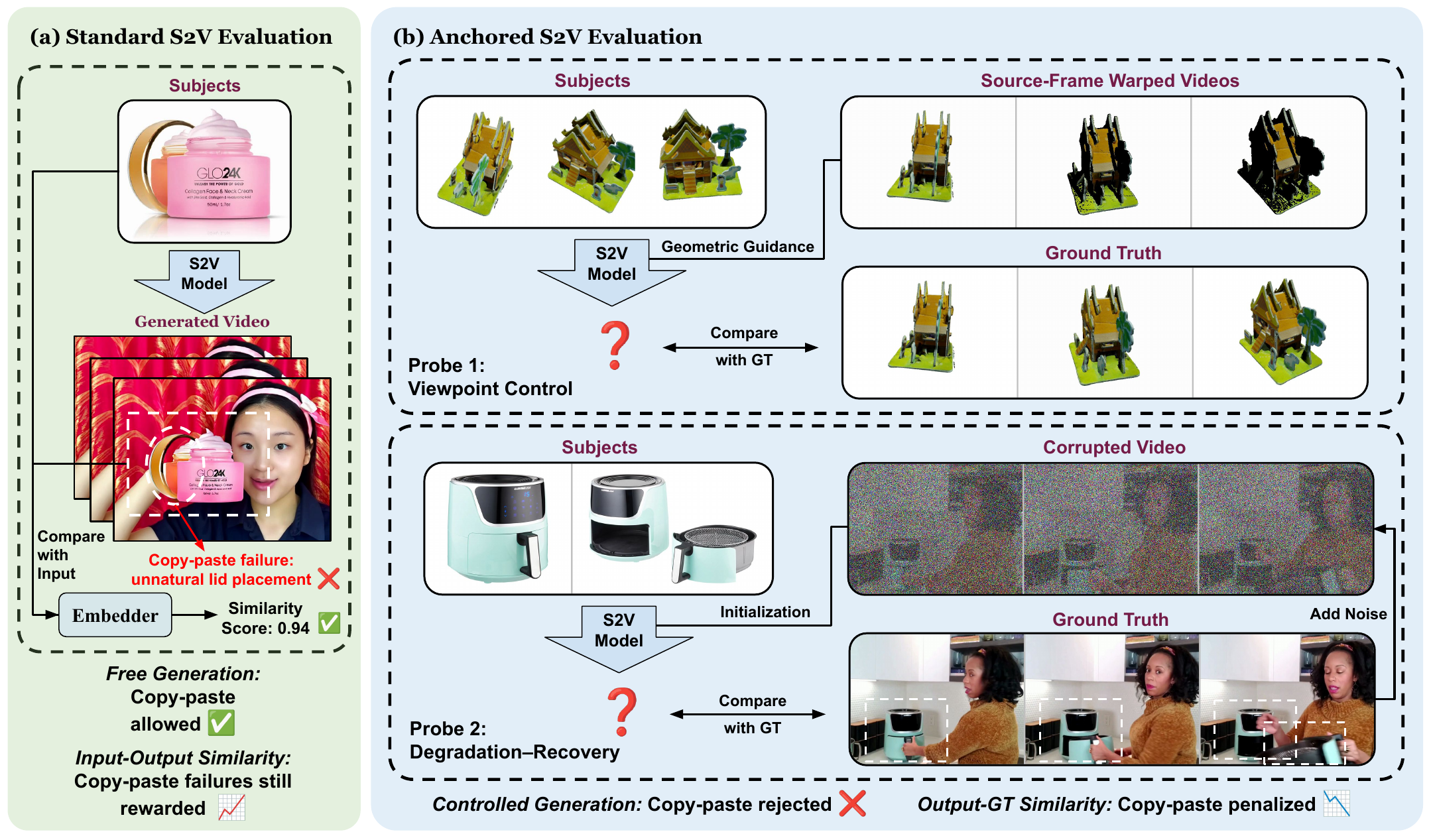}
\caption{(a)~Conventional evaluation uses free generation and input-similarity scoring. Models can copy-paste reference content, even producing implausible results (e.g., unnatural lid placement), yet still receive high similarity scores as the metric rewards surface-level copying rather than subject understanding. (b)~Anchored evaluation controls generation via two probes and compares against GT, structurally preventing copy-pasting: viewpoint control specifies target geometry via depth-based warping, and the model must complete occluded regions using subject appearance from the references; degradation--recovery corrupts a held-out video and recovers it conditioned on the references.}
\vspace{-10pt}
\label{fig:teaser}
\end{figure*}

Recent advances in diffusion models~\citep{ho2020denoising,dit,rombach2022high} have driven remarkable progress in video generation~\citep{Make-a-video,animatediff,cogvideox,hunyuanvideo,wan21}, enabling modern systems to produce high-fidelity, temporally coherent sequences from textual prompts alone.
Building on these advances, \emph{subject-driven video generation}, the task of producing videos that faithfully incorporate a user-provided reference subject, has emerged as a critical capability for applications such as product advertising, personalized content creation, and brand marketing~\citep{phantom,consisid,conceptmaster,vace,skyreelsa2}.

Current evaluation of subject-driven models relies on \emph{free generation} (\cref{fig:teaser}a): a model generates a video from random noise conditioned on reference images, and subject fidelity is scored by CLIP~\citep{clip} or DINOv2~\citep{oquab2023dinov2} similarity to the input references, or by VLM-based scoring~\citep{dreambench}.
This protocol has a fundamental blind spot.
Because the model is free to choose an easy generation trajectory, it can obtain high scores by reproducing visible cues from the reference (textures, viewpoints, or poses) while avoiding transformations that would test whether the subject is preserved beyond the conditioning input.
Moreover, reference-similarity metrics actively reward this shortcut, making it difficult to distinguish a model that leverages subject conditioning to infer unseen content from one that merely copies what it was given.
Our argument is about this evaluation shortcut, not a preference for diversity for its own sake: reproducing a reference can be appropriate when it matches the requested target, while a capable model should also generalize when the target requires unseen views or states.
A meaningful evaluation should instead test whether models can go beyond copying and generalize the subject to unseen conditions (\cref{fig:teaser}b), e.g., inferring how the subject appears from a novel viewpoint or behaves in an unseen state.

Toward this goal, we design controlled evaluation settings, collectively referred to as \textbf{anchored evaluation}, in which generation is anchored to held-out targets of the same subject rather than left entirely free. Each setting defines a recovery or synthesis task using a held-out target, while withholding the target observation from the model's subject conditioning.
The target provides structural or geometric guidance (a noisy starting point in degradation--recovery, or warped geometry in viewpoint control), but subject-specific identity and appearance must come entirely from the conditioning references.
Success in our anchored evaluation therefore requires the model to use the references to infer content not directly shown in its conditioning, rather than simply copying what it was given.
Moreover, because fidelity is measured against the held-out ground truth rather than the input references, metrics no longer reward surface-level copying.

We instantiate this through two probes (\cref{fig:teaser}b):
(1)~\emph{viewpoint control}, which warps a held-out target frame to a novel viewpoint via depth-based warping, producing visible geometry but leaving occluded regions that the model must fill in by drawing on subject appearance from the references; and
(2)~\emph{degradation--recovery}, which adds calibrated noise to a held-out video of the subject in a real-world context that differs from the reference images, then asks the model to denoise it conditioned on those references, testing whether it can generalize subject identity across contexts.
Both probes slot into the standard inference flow of flow-matching models without retraining and apply uniformly across architectures.
Together, they test complementary axes: viewpoint control tests spatial generalization across views, while degradation--recovery tests cross-context generalization across environments and states.

We construct evaluation data from real-world product listings and scanned 3D objects, and evaluate across seven diverse, strong models.
Anchored evaluation reveals a substantially different picture from conventional metrics.
First, rankings shift dramatically: VACE-14B ranks first under conventional evaluation (0.739) while Phantom-14B places last (0.633), yet under anchored evaluation Phantom-14B leads.
Second, conventional scores strongly correlate with the improvement gained from providing ground-truth references (Pearson $r = 0.91$), suggesting they largely measure copying ability rather than subject understanding.
Third, fine-grained analysis reveals distinct failure modes: copy-paste models drift at intermediate viewpoints even with ground-truth references, while no model adaptively balances copying and generalization.
A 2,000-comparison human study validates this diagnosis: human preferences correlate strongly with anchored scores across both probes and reference settings (Spearman $\rho{=}0.82$--$0.91$), but only weakly with conventional scores ($\rho{=}0.14$).
These results suggest conventional evaluation overstates progress by rewarding copying, while anchored evaluation offers a more diagnostic view of subject conditioning.

In summary, this work makes three contributions:
(1)~we identify the reference-copying blind spot in current evaluation and design anchored evaluation, two training-free probes that test subject conditioning beyond surface copying;
(2)~ we conduct a systematic study across seven diverse models, validated by 2,000 human comparisons and multiple alternative metrics, showing that rankings shift substantially under anchored evaluation; and
(3)~we analyze failure modes across architectures, revealing when and how models fall back on reference copying.

\section{Related Work}
\label{sec:related}

\subsection{Subject-Driven Video Generation}
\label{sec:related-method}
Subject-driven video generation sits between general-purpose text-to-video (T2V) foundation models, which range from early inflated 3D U-Nets~\citep{Make-a-video,animatediff,blattmann2023stable} to large DiT-based open-source~\citep{dit,cogvideox,hunyuanvideo,wan21,opensora,vchitect20} and closed-source~\citep{movie_gen,sora2,veo3,seedance2,kling3,runwaygen4} systems but mostly offer no native subject-conditioning mechanism, and subject-driven image (S2I) generation, where DreamBooth~\citep{Dreambooth}, Textual Inversion~\citep{textual_inversion}, and IP-Adapter~\citep{Ip-adapter} established the per-subject-optimization and feed-forward templates and a long tail of follow-up extends them to encoder-based concept embeddings, identity-preserving face generation, and unified or controllable image-generation backbones~\citep{Elite,blip_diffusion,photomaker,pulid,instantid,controlnet,sdxl,Imagen,DALLE2}.
S2V inherits both this lineage and its CLIP/DINO~\citep{clip,oquab2023dinov2} reference-similarity evaluation protocol: early S2V followed per-subject optimization~\citep{magic-me,wang2024customvideo,chen2024disenstudio,motionbooth,dreamvideo}; a second wave introduced feed-forward conditioning focused on facial identity and human animation~\citep{ID-Animator,consisid,fantasyid,concat-id,animateanyone,animateanyone2,magicdance}; and a current wave of DiT-based backbones, with Phantom~\citep{phantom} and VACE~\citep{vace} as representatives, enables open-set, multi-subject S2V at scale~\citep{skyreelsa2,conceptmaster,magref,cinema,hunyuancustom,msrvttpersonlize,videomaker,movieweaver,zhang2025kaleido,bindweave,saber,reward_guided_i2v,ingredients,videobooth}.
Despite this methodological diversity, every S2V system is trained and benchmarked under \emph{few-image conditioning} and \emph{reference-similarity} evaluation, which we argue in \cref{sec:related-eval} systematically rewards appearance-copying over generalization to novel views and contexts.

\subsection{Evaluation of Subject-Driven Video Generation}
\label{sec:related-eval}
General T2V evaluation spans distributional metrics~\citep{fvd}, multi-axis benchmarks~\citep{Vbench,vbench++,vbench20}, learned reward models~\citep{videoscore,T2VScore}, improved distributional measures~\citep{beyond_fvd}, compositional benchmarks~\citep{evalcrafter,FETV,t2vcompbench,chronomagicbench}, and VLM/MLLM judges~\citep{dreambench,mllm_judge}, but none target subject conditioning.
For subject fidelity, the DreamBooth protocol~\citep{Dreambooth} (CLIP-I, DINO, CLIP-T) remains the de facto standard across S2I and S2V alike; DreamBench++~\citep{dreambench} adds VLM-based scoring with multi-subject and story-level extensions~\citep{multihuman_testbench,refvnli,identitystory}.
S2V-specific evaluation remains limited: face-centric protocols tied to identity-preserving methods~\citep{consisid,ID-Animator,face_consistency_bench}, open-domain evaluations reusing VBench's subject-consistency dimension~\citep{vace,skyreelsa2}, a multi-subject protocol~\citep{msrvttpersonlize}, and OpenS2V-Nexus~\citep{yuan2025opens2v} with retrieval-based scoring~\citep{gme}.
Three structural limitations cut across this landscape: every protocol uses generated-to-reference similarity as the dominant subject signal, incentivizing copying; conditioning is universally sparse, making it impossible to tell whether a model lacks subject information or fails to use it; and no protocol evaluates against \emph{held-out} observations, so high scores cannot be distinguished from genuine understanding.
While novel-view synthesis benchmarks~\citep{mipnerf360,tanksandtemples} also score against held-out targets, they condition on dense multi-view captures and test geometric reconstruction; anchored evaluation instead conditions on $\le 3$ sparse references and tests subject-identity transfer.
We address these gaps by shifting the evaluation target from the conditioning input to held-out observations of the same subject.

\section{Anchored Evaluation}
\label{sec:method}

This section presents anchored evaluation, a set of controlled probes for subject-driven video generation.
We introduce two training-free procedures that test whether models can recover held-out targets by leveraging subject conditioning, rather than copying visible cues from the references.
Both operate as plug-and-play modifications to the inference loop of flow-matching diffusion models, requiring no retraining or architectural changes.
We first review the necessary preliminaries~(\cref{sec:prelim}), then describe the two probes: degradation--recovery~(\cref{sec:degrade_recover}) and viewpoint control~(\cref{sec:novel_view}).

\subsection{Preliminaries}
\label{sec:prelim}

We build on latent video diffusion models.
A pretrained video autoencoder compresses an RGB video $\mathbf{x}$ into a latent $\mathbf{z} = \mathcal{E}(\mathbf{x})$.
Flow matching~\citep{lipmanflow} constructs a noisy latent via linear interpolation:
\begin{equation}
\label{eq:flow_forward}
\mathbf{z}_t = (1 - \sigma_t)\,\mathbf{z}_0 + \sigma_t\,\boldsymbol{\epsilon}, \quad \boldsymbol{\epsilon} \sim \mathcal{N}(\mathbf{0}, \mathbf{I}),
\end{equation}
where $\sigma_t$ is the noise level at diffusion time $t$. We distinguish diffusion time from the sampling-step index: an $S$-step sampling schedule follows noise times $t_0 > t_1 > \cdots > t_S = 0$.
A denoising network $\boldsymbol{\epsilon}_\theta(\mathbf{z}_t, t, \mathbf{c})$ predicts the flow velocity $\boldsymbol{\epsilon} - \mathbf{z}_0$, yielding a clean estimate $\hat{\mathbf{z}}_0 = \mathbf{z}_t - \sigma_t\,\boldsymbol{\epsilon}_\theta$.
During inference, classifier-free guidance~\citep{CFG} interpolates between conditional and unconditional predictions to improve fidelity, and a scheduler iteratively denoises from $\mathbf{z}_{t_0}$ to $\mathbf{z}_0$, which is decoded as $\hat{\mathbf{x}} = \mathcal{D}(\mathbf{z}_0)$.
Subject-driven models condition on both a text prompt $\mathbf{c}_\text{text}$ and reference images $\mathbf{c}_\text{ref}$, injected via model-specific mechanisms (e.g., cross-attention, latent concatenation, or leading-frame encoding~\citep{phantom,vace,skyreelsa2}).

\subsection{Degradation--Recovery}
\label{sec:degrade_recover}

Degradation--recovery tests whether models can reconstruct a held-out target video of the same subject using their subject conditioning.
The target videos are real-world demonstration clips depicting the subject undergoing state changes (e.g., being opened, assembled, or operated), while the reference images are product photos captured in entirely different contexts (different environment, lighting, and camera perspective).
Because references and target share the same subject but differ in everything else, recovery cannot succeed by pixel copying.
Moreover, the references depict the subject's states statically, while the target captures the dynamic transition between them; faithful reconstruction thus requires both cross-context generalization of subject appearance and inference of state dynamics from the reference conditioning.

\paragraph{Formulation.}
Given a target video $\mathbf{V}_\text{tgt}$ and subject condition $\mathbf{c} = (\mathbf{c}_\text{text}, \mathbf{c}_\text{ref})$, we first encode the target into the latent space: $\mathbf{z}_\text{tgt} = \mathcal{E}(\mathbf{V}_\text{tgt})$.
We then construct a noisy latent at a chosen sampling-step index $G$ ($0 < G < S$), writing $\sigma_G \equiv \sigma_{t_G}$ for the noise level at that index:
\begin{equation}
\label{eq:add_noise}
\mathbf{z}_{t_G} = (1 - \sigma_G)\,\mathbf{z}_\text{tgt} + \sigma_G\,\boldsymbol{\epsilon}, \quad \boldsymbol{\epsilon} \sim \mathcal{N}(\mathbf{0}, \mathbf{I}).
\end{equation}
For models that encode reference images as leading latent frames, these frames must remain as pure noise $\sigma_G\,\boldsymbol{\epsilon}$ to match the model's expected input distribution, while the noised target is injected only into the video region of the latent tensor.
Starting from $\mathbf{z}_{t_G}$, we run the standard denoising loop for the remaining $S - G$ steps with subject conditioning $\mathbf{c}$, producing the reconstructed video $\hat{\mathbf{V}} = \mathcal{D}(\mathbf{z}_0)$.

The guide step $G$ controls a fidelity--flexibility tradeoff.
A smaller $G$ (earlier in the schedule) corresponds to higher noise $\sigma_G$, granting the model more freedom to deviate from the target; a larger $G$ (later in the schedule) yields lower noise, constraining the output to closely follow the original content.
This is the sole hyperparameter.

\paragraph{Data.}
We source evaluation data from Amazon product pages, which naturally provide both product images (reference) and demonstration videos (target) for the same item.
We curate 100 reference--video pairs spanning 67 product categories, selecting video clips that exhibit meaningful state changes (e.g., a backpack being opened, an appliance being operated) whose outcomes can be inferred from the accompanying product images.
Each sample includes at least three reference images from different angles or configurations; details of the filtering pipeline are in Appendix \cref{sec:dr_data_details}.

\subsection{Viewpoint Control}
\label{sec:novel_view}

The second probe tests cross-view identity preservation by generating novel viewpoints of the subject that are absent from the reference set.
Given a target viewpoint and a set of reference images from other views, we guide the generation process to follow the specified camera trajectory.
Depth-based warping supplies pixel content for regions visible from the source view, but self-occluded regions appear as holes that the model must fill by inferring the subject's appearance from the references, directly testing whether the model can synthesize unseen surfaces from multi-view conditioning.

\paragraph{Stage 1: Reference Frame Warping.}
We first produce geometric guidance by warping a source image to each target viewpoint via point-cloud reprojection.
Given the source image $\mathbf{I}_\text{src}$, its depth map $\mathbf{D}_\text{src}$ (from ground-truth 3D assets), and camera intrinsics $\mathbf{K}$, we unproject the source pixels into a 3D point cloud and render it from each target camera pose $\mathbf{P}_t$.
Regions of the target view that are occluded or not visible from the source naturally become holes in the rendering.
This yields a warped frame $\mathbf{W}_t$ and a validity mask $\mathbf{M}_t \in [0, 1]^{H \times W}$:
\begin{equation}
\label{eq:warp}
(\mathbf{W}_t,\;\mathbf{M}_t) = \mathcal{W}(\mathbf{I}_\text{src},\;\mathbf{D}_\text{src},\;\mathbf{P}_\text{src},\;\mathbf{P}_t), \quad t = 1, \ldots, T,
\end{equation}
where $\mathbf{M}_t \approx 1$ in regions with valid reprojected content and $\mathbf{M}_t \approx 0$ in holes.
Masks are post-processed with morphological closing and edge softening, and capped at $M_\text{cap}$ (e.g., $0.95$) to allow partial freedom even in visible regions.

\paragraph{Stage 2: Camera-Guided Denoising.}
Following WorldForge~\citep{worldforge}, we inject the warped frames into the denoising loop as pixel-space guidance during the first $G$ of $S$ total steps, with $G=S$ in our viewpoint-control experiments.
At each guided sampling step $s=0,\ldots,G-1$, we first obtain the clean latent estimate $\hat{\mathbf{z}}_0$ from the current noisy state $\mathbf{z}_{t_s}$ using classifier-free guidance, then decode it to pixel space as $\hat{\mathbf{I}}_s = \mathcal{D}(\hat{\mathbf{z}}_0)$.
The decoded frame is blended with the warped reference via the validity mask:
\begin{equation}
\label{eq:blend}
\mathbf{I}_\text{fused} = \mathbf{W}_t \odot \mathbf{M}_t + \hat{\mathbf{I}}_s \odot (1 - \mathbf{M}_t),
\end{equation}
where $\mathbf{M}_t$ acts as a soft spatial selector between the warped geometry and the model's own prediction.
The fused frame is re-encoded and used to correct the velocity prediction: $\boldsymbol{\epsilon}' = (\mathbf{z}_{t_s} - \mathcal{E}(\mathbf{I}_\text{fused})) / \sigma_{t_s}$.
To strengthen adherence to the geometric guidance, we apply directional similarity guidance (DSG)~\citep{worldforge}, which pushes $\boldsymbol{\epsilon}'$ further along the component orthogonal to the uncorrected prediction $\boldsymbol{\epsilon}_\theta$:
\begin{equation}
\label{eq:dsg}
\hat{\boldsymbol{\epsilon}} = \boldsymbol{\epsilon}' + \omega \sin\theta\,\bigl(\boldsymbol{\epsilon}' - \|\boldsymbol{\epsilon}'\|/\|\boldsymbol{\epsilon}_\theta\| \cdot \cos\theta\,\boldsymbol{\epsilon}_\theta\bigr),
\end{equation}
where $\theta$ is the angle between $\boldsymbol{\epsilon}'$ and $\boldsymbol{\epsilon}_\theta$, and $\omega$ controls the guidance strength.
When $G<S$, the remaining $S - G$ steps proceed with standard denoising.
Throughout the entire process, the model is simultaneously conditioned on the subject references $\mathbf{c}_\text{ref}$, which provide identity information that the geometric warping alone cannot supply.

\paragraph{Data.}
We source 3D objects from OmniObject3D~\citep{omniobject3d}, which provides high-quality scanned meshes with ground-truth depth maps and multi-view Blender renders.
We select 100 objects from 100 categories based on DINOv2~\citep{oquab2023dinov2} view disparity scores, favoring objects with sufficient cross-view appearance variation to distinguish model capabilities, while filtering out near-symmetric or textureless objects where viewpoint control is trivial.
For each object, we select a source--target view pair with moderate angular separation (avoiding both near-identity and extreme occlusion regimes) and generate 49-frame warped sequences by rendering the source point cloud along the interpolated camera trajectory.
Three reference images are sampled from held-out viewpoints not on the warping trajectory: two near the target view and one near the source view, covering the range of appearances from source to target.
Details are in Appendix \cref{sec:vp_data_details}.

\begin{figure*}[t]
\centering
\includegraphics[width=\textwidth]{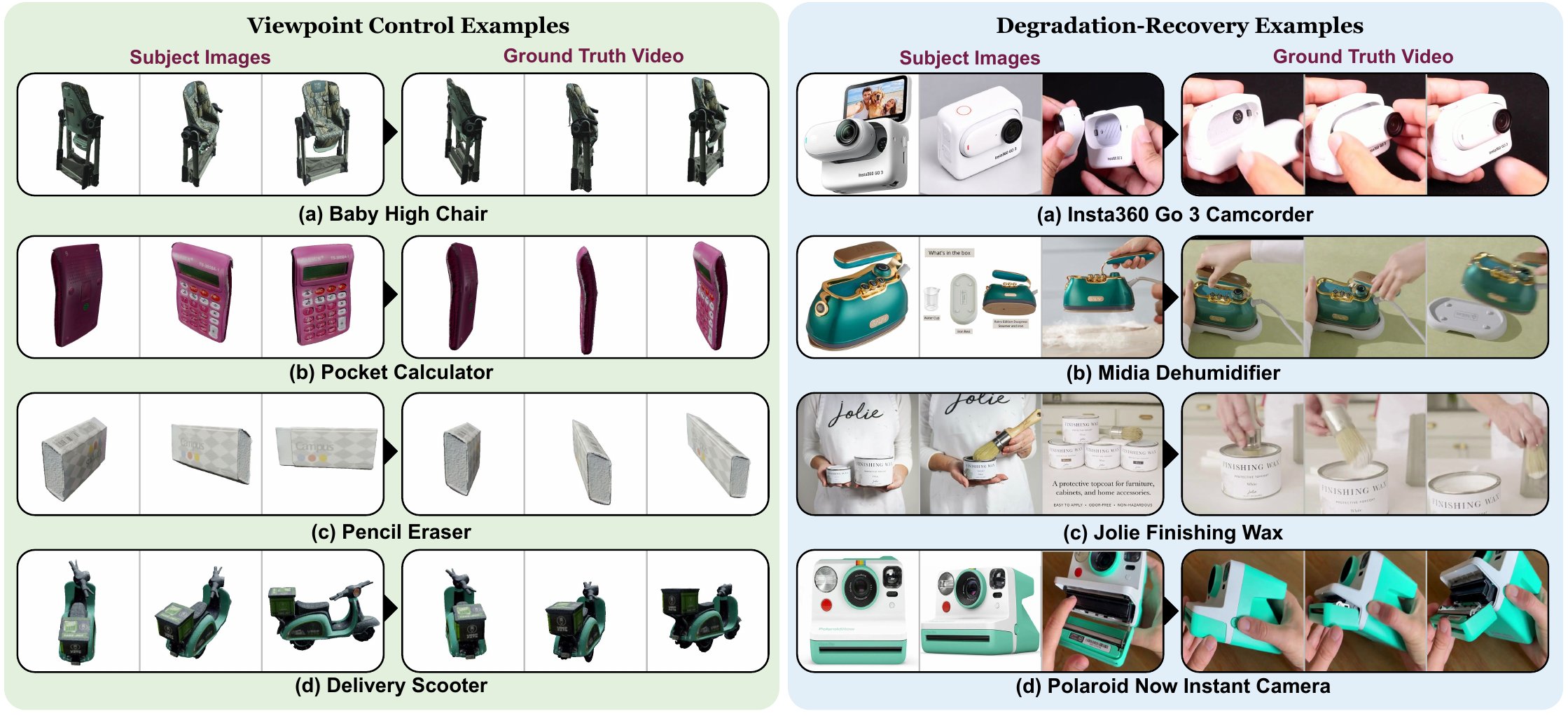}
\caption{Example evaluation samples. \textbf{Left:} Viewpoint control: subject images from held-out reference views and ground-truth video frames showing the target viewpoint trajectory. \textbf{Right:} Degradation--recovery: reference product images and ground-truth video frames showing the subject undergoing real-world state changes.}
\label{fig:data_samples}
\end{figure*}

\subsection{Evaluation Metrics}
\label{sec:eval_metrics}

Both probes measure how well the generated output preserves the ground-truth subject by comparing against held-out target frames.
We report two similarity metrics: DINOv2~\citep{oquab2023dinov2} cosine similarity, which captures semantic-level subject fidelity, and GME~\citep{gme}, a multimodal-retrieval-based metric adopted from OpenS2V-Nexus~\citep{yuan2025opens2v}.
Our primary metric is the average of DINOv2 and GME scores.

For degradation--recovery, we evaluate subject-level fidelity on annotated key frames: for each target video, we annotate bounding boxes around the subject and select key frames exhibiting distinctive states, then compute DINOv2 and GME similarity between cropped subject patches in the generated and ground-truth videos.
For viewpoint control, we crop both generated and ground-truth target frames using the object mask so that the subject is dominant in the comparison, then compute per-frame DINOv2 and GME similarity averaged across sampled frames along the camera trajectory.

\section{Experiments}
\label{sec:experiments}

\subsection{Evaluation Setup}
\label{sec:eval_setup}

\paragraph{Models.}
We evaluate seven strong subject-driven video generation models that support multi-reference conditioning:
Phantom 1.3B and 14B~\citep{phantom},
VACE 1.3B and 14B~\citep{vace},
MAGREF-14B~\citep{magref},
SkyReels-A2-14B~\citep{skyreelsa2},
and Kaleido-14B~\citep{zhang2025kaleido},
spanning different conditioning mechanisms (cross-attention injection, leading-frame concatenation, masked guidance, etc.) and model scales (1.3B and 14B parameters).

\paragraph{Reference Settings.}
We evaluate each model under two reference settings.
In the \emph{canonical} setting, we provide the three reference images selected during data curation (\cref{sec:degrade_recover,sec:novel_view}), which depict the subject in different views or contexts from the target; success therefore requires the model to infer held-out content from the references rather than copy them.
In the \emph{oracle} setting, references are sampled directly from the ground truth: for viewpoint control, we provide the target view itself; for degradation--recovery, we uniformly sample three frames from the GT video.
Because oracle references contain the target appearance, this setting upper-bounds canonical performance and reveals how much each model benefits when direct copying is possible.

\paragraph{Generation and Evaluation Protocol.}
All models generate 49 frames at $480 \times 832$ resolution (or the nearest resolution each model supports) with official default hyperparameters.
Degradation--recovery uses a unified noise level $\sigma_G = 0.972$; ablations over $\sigma_G$ are provided in Appendix \cref{sec:sigma_ablation}.
For viewpoint control, we apply pixel-space fusion with DSG guidance ($\omega = 1.8$) throughout all denoising steps and evaluate on frames $f_{21}$, $f_{33}$, and $f_{49}$, as early frames are dominated by warped content and show little variation across models; ablations over $\omega$ are provided in Appendix \cref{sec:omega_ablation}.

\paragraph{Conventional Free-Generation Baseline.}
As a reference for conventional (non-anchored) evaluation, we run each model in standard free-generation mode on the same items used in the anchored probes.
For viewpoint control, the target view serves as the sole reference image; for degradation--recovery, we use the first reference image from the curated set.
Following~\citet{yuan2025opens2v}, we detect and crop the subject in each generated frame, then compute DINO and GME similarity against the input reference; the reported score is the average across both probes.

\begin{table*}[t]
\centering
\caption{\textbf{Main results under anchored evaluation.} For conventional free generation on the same items, we report DINO, GME, and their average separately. For each probe, we report results under a \emph{canonical} setting (references differ from the target) and an \emph{oracle} setting (references are from the ground truth). $\textit{avg} = (\text{DINO} + \text{GME})/2$.}
\label{tab:main}
\setlength{\tabcolsep}{3pt}
\renewcommand{\arraystretch}{1.05}
\small
\resizebox{\textwidth}{!}{\begin{tabular}{l ccc | ccc ccc c cccc cccc}
\toprule
\multirow{3}{*}{Model} & \multicolumn{3}{c}{\textbf{Conventional}} & \multicolumn{6}{c}{\textbf{Degradation--Recovery (DR)}} & & \multicolumn{8}{c}{\textbf{Viewpoint Control (VC)}} \\
\cmidrule(lr){2-4} \cmidrule(lr){5-10} \cmidrule(lr){12-19}
& \multicolumn{3}{c}{\textbf{Free Gen.}} & \multicolumn{3}{c}{\emph{canonical}} & \multicolumn{3}{c}{\emph{oracle}} & & \multicolumn{4}{c}{\emph{canonical}} & \multicolumn{4}{c}{\emph{oracle}} \\
\cmidrule(lr){2-4} \cmidrule(lr){5-7} \cmidrule(lr){8-10} \cmidrule(lr){12-15} \cmidrule(lr){16-19}
& DINO$\uparrow$ & GME$\uparrow$ & avg$\uparrow$ & DINO$\uparrow$ & GME$\uparrow$ & avg$\uparrow$ & DINO$\uparrow$ & GME$\uparrow$ & avg$\uparrow$ & & PSNR$\uparrow$ & DINO$\uparrow$ & GME$\uparrow$ & avg$\uparrow$ & PSNR$\uparrow$ & DINO$\uparrow$ & GME$\uparrow$ & avg$\uparrow$ \\
\midrule
Phantom-1.3B    & 0.654 & 0.716 & 0.685 & 0.708 & 0.819 & 0.764 & 0.855 & 0.903 & 0.879 & & 21.64 & 0.838 & 0.855 & 0.846 & \textbf{22.61} & 0.856 & 0.864 & 0.860 \\
Phantom-14B     & 0.609 & 0.657 & 0.633 & \textbf{0.726} & \textbf{0.842} & \textbf{0.784} & 0.867 & 0.896 & 0.881 & & \textbf{21.76} & \textbf{0.843} & \textbf{0.856} & \textbf{0.849} & 22.51 & 0.850 & 0.866 & 0.858 \\
VACE-1.3B       & 0.611 & 0.721 & 0.666 & 0.642 & 0.798 & 0.720 & 0.764 & 0.861 & 0.812 & & 20.17 & 0.805 & 0.837 & 0.821 & 20.66 & 0.851 & 0.865 & 0.858 \\
VACE-14B        & \textbf{0.703} & \textbf{0.775} & \textbf{0.739} & 0.693 & 0.831 & 0.762 & \textbf{0.893} & \textbf{0.948} & \textbf{0.920} & & 20.26 & 0.791 & 0.841 & 0.816 & 22.53 & \textbf{0.857} & \textbf{0.873} & \textbf{0.865} \\
SkyReels-A2-14B & 0.693 & 0.763 & 0.728 & 0.660 & 0.784 & 0.722 & 0.844 & 0.890 & 0.867 & & 19.44 & 0.806 & 0.831 & 0.819 & 20.64 & 0.843 & 0.849 & 0.846 \\
Kaleido-14B     & 0.647 & 0.741 & 0.694 & 0.669 & 0.809 & 0.739 & 0.781 & 0.873 & 0.827 & & 20.98 & 0.786 & 0.823 & 0.805 & 22.10 & 0.847 & 0.851 & 0.849 \\
MAGREF-14B      & 0.601 & 0.687 & 0.644 & 0.673 & 0.818 & 0.745 & 0.826 & 0.898 & 0.862 & & 20.54 & 0.784 & 0.827 & 0.806 & 20.86 & 0.794 & 0.832 & 0.813 \\
\bottomrule
\end{tabular}}
\end{table*}

\subsection{Conventional vs.\ Anchored Evaluation}
\label{sec:conv_vs_anchored}

\begin{figure}[t]
\centering
\begin{minipage}[t]{0.45\linewidth}
\centering
\includegraphics[width=0.95\linewidth]{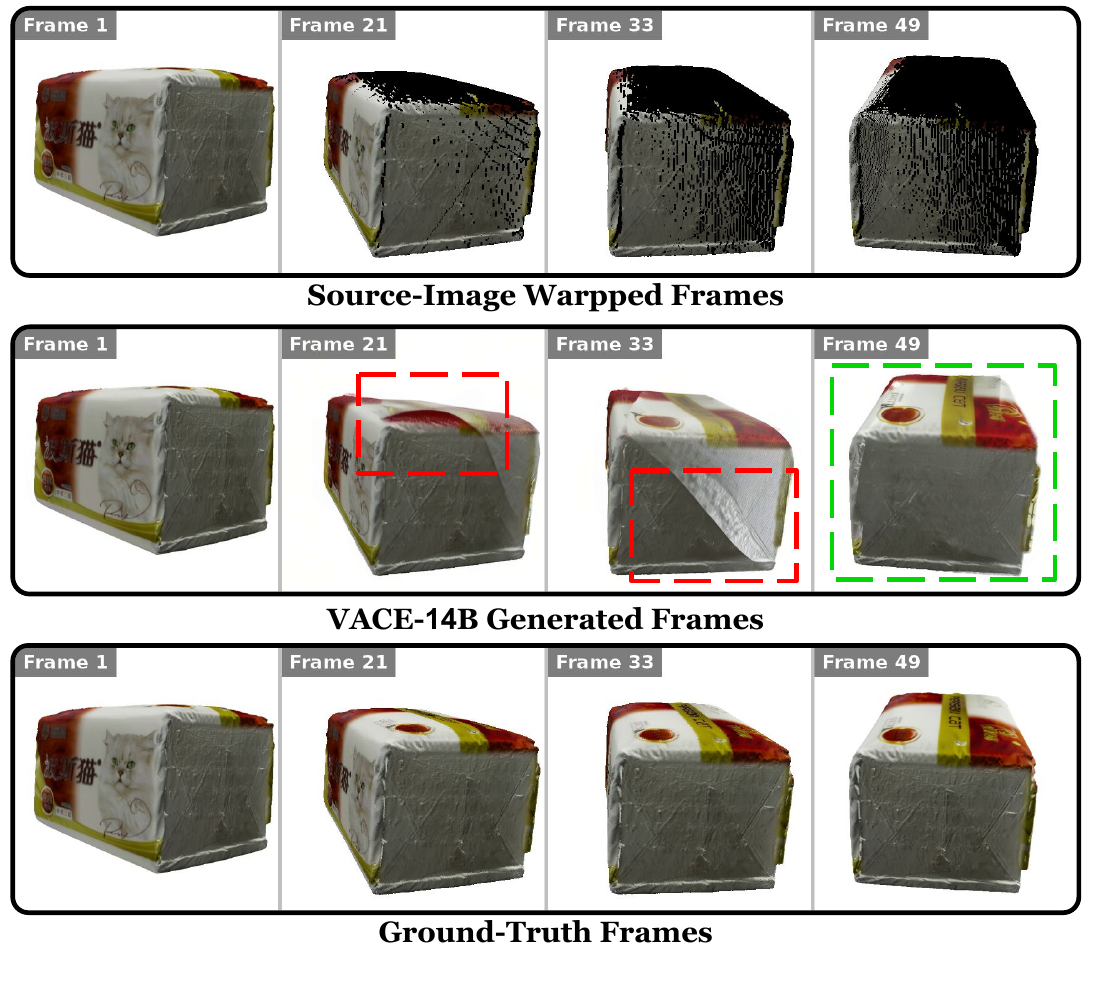}
\caption{VACE-14B drifting example at intermediate viewpoints with oracle references (frame $f_{49}$ of the GT). Within invisible regions of the warped frames, the model copies the last frame but produces inconsistent content at intermediate frames (red boxes).}
\label{fig:view_inconsistency}
\end{minipage}
\hfill
\begin{minipage}[t]{0.52\linewidth}
\centering
\includegraphics[width=\linewidth]{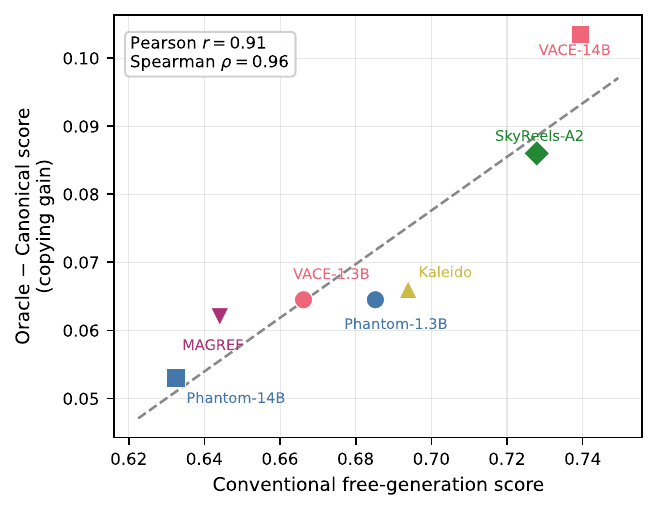}
\vspace{-10pt}
\caption{Models with higher conventional free-generation scores also improve more when switching from canonical to oracle references (Pearson $r = 0.91$, Spearman $\rho = 0.96$), indicating that conventional evaluation rewards reference-copying ability rather than subject understanding.}
\label{fig:correlation}
\end{minipage}
\end{figure}

We compare conventional free-generation rankings and scores with anchored results under canonical and oracle settings in \Cref{tab:main,fig:correlation}. Qualitative comparisons under both canonical and oracle settings are also provided in Appendix \cref{sec:qualitative}.

\paragraph{Conventional rankings shift under anchored evaluation.}
\Cref{tab:main} reveals notable inconsistencies between conventional and anchored rankings.
Under anchored evaluation, the top-ranked model differs across settings: Phantom-14B ranks first under canonical references, while VACE-14B ranks first under oracle references.
Yet under conventional free-generation evaluation, their ordering is entirely reversed: VACE-14B places first (0.739) and Phantom-14B last (0.633).
Conventional scores therefore cannot reliably indicate how well a model understands its subject.

\paragraph{Conventional scores strongly correlate with copying gain.}
 
We define \emph{copying gain} as $\Delta_{\mathrm{copy}}=S_{\mathrm{oracle}}-S_{\mathrm{canonical}}$ for similarity metrics (with the sign reversed for distance metrics). This is a matched contrast: the target, anchored trajectory, guidance, samples, and metric are fixed, and only the subject-reference condition changes from held-out to target-matched. A raw oracle score conflates the ability to follow the imposed trajectory and reconstruct the target with the ability to exploit target-matched references. Subtracting canonical performance removes much of this shared task difficulty and isolates the marginal benefit of making direct reference reuse possible. Conventional free-generation scores strongly correlate with this gain (Pearson $r{=}0.91$, Spearman $\rho{=}0.96$; \Cref{fig:correlation}), indicating that reference-similarity evaluation rewards the same copy-compatible behavior. This does not imply that copying is intrinsically undesirable or that every high oracle score is caused by copying; it shows that conventional scores cannot separate copying from held-out subject generalization.

\paragraph{Summary.}
Both findings indicate that conventional scores largely reflect reference-copying ability. By controlling the reference content, anchored evaluation disentangles copying from generalization and reveals capabilities that conventional evaluation cannot surface.

\subsection{Human Validation and Metric Robustness}
\label{sec:human_validation}

We validate the automatic rankings with 2,000 pairwise human comparisons: 400 comparisons for conventional free generation and for each canonical/oracle setting of the two anchored probes. For each comparison, raters view anonymized, order-randomized outputs generated from the same sample and prompt; ties are allowed. Conventional outputs are judged for subject preservation and overall video quality, whereas anchored outputs are additionally compared against the held-out ground truth. We aggregate pairwise outcomes with a Bradley--Terry model, counting ties as 0.5.

\begin{table}[H]
\centering
\caption{Correlation between automatic model rankings and human preferences across seven models. Anchored evaluation aligns strongly with human judgments, whereas conventional evaluation does not.}
\label{tab:human_validation}
\small
\setlength{\tabcolsep}{7pt}
\begin{tabular}{lcc}
\toprule
Setting & Pearson $r$ & Spearman $\rho$ \\
\midrule
Conventional free generation & $-0.12$ & $0.14$ \\
VC canonical & $0.77$ & $0.82$ \\
VC oracle & $0.82$ & $0.87$ \\
DR canonical & $0.79$ & $0.91$ \\
DR oracle & $0.81$ & $0.85$ \\
\bottomrule
\end{tabular}
\end{table}

A concrete case is Phantom-14B versus VACE-14B: human judgments find their subject fidelity broadly comparable, whereas conventional reference similarity ranks Phantom-14B substantially lower because it tends to synthesize novel views rather than reproduce the input appearance. Anchored evaluation instead scores both models against the requested held-out target, so novelty is neither rewarded nor penalized by itself.

The conclusion is also robust to alternative metrics. CLIP-based model rankings correlate strongly with the DINO/GME rankings across anchored settings (Spearman $\rho{=}0.81$--$1.00$). Replacing identity similarity with ID-Sim preserves the association between conventional performance and copying gain ($r{=}0.84$, $\rho{=}0.96$), while the oracle--canonical gain in VC PSNR is likewise correlated with conventional performance ($r{=}0.84$, $\rho{=}0.89$). The conventional DINO and GME components independently preserve the original average ranking ($\rho{=}0.93$ and $0.96$) and both correlate with copying gain. Full protocols, per-model results, and additional evaluations are provided in Appendix~\cref{sec:human_details,sec:metric_robustness,sec:hdtf,sec:temporal,sec:closed_source,sec:opens2v_comparison}.

\subsection{Fine-Grained Analysis}
\label{sec:ref_utilization}

\begin{figure*}[t]
\centering
\includegraphics[width=\textwidth]{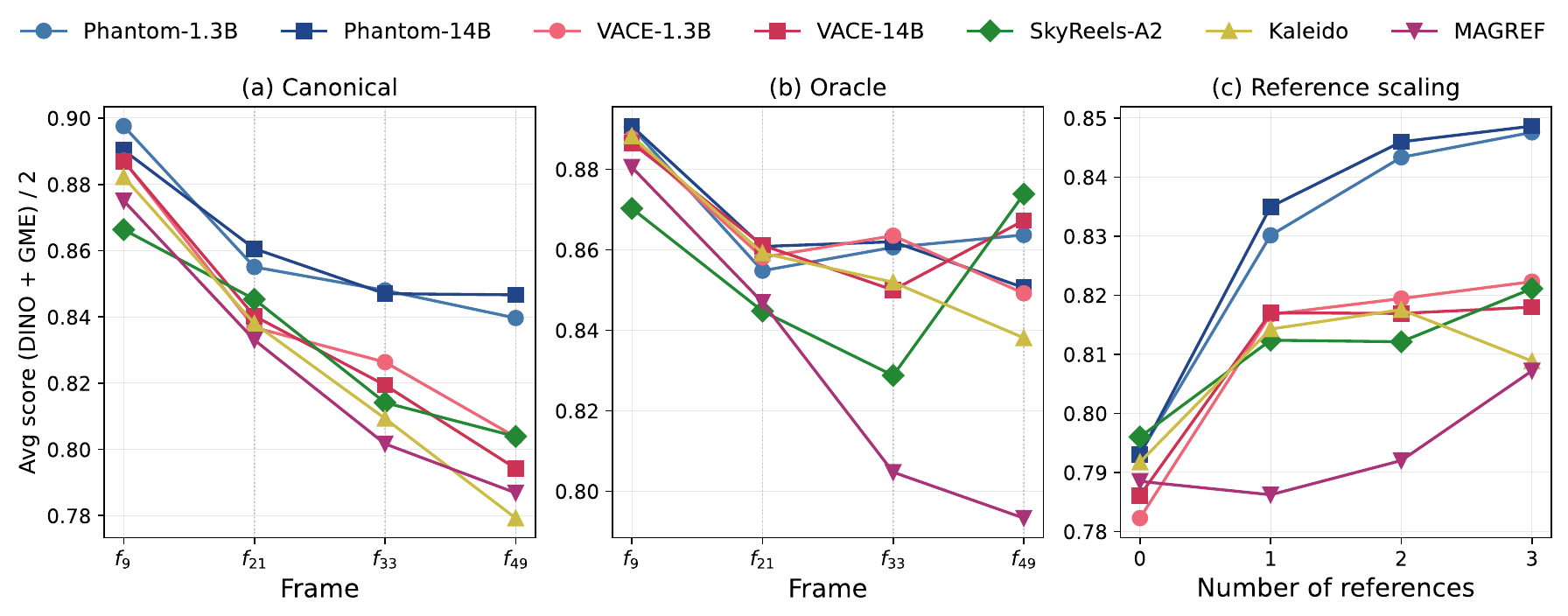}
\vspace{-15pt}
\caption{Viewpoint control analysis. \textbf{(a)} Canonical references: Phantom models show smaller declines in identity scores across frames, while VACE, Kaleido, and MAGREF degrade sharply toward the target view. \textbf{(b)} Oracle references: VACE-14B and SkyReels-A2 exhibit V-shaped curves consistent with copy-paste behavior. \textbf{(c)} Reference scaling: Phantom models improve consistently with more references, while several models plateau or decline.}
\label{fig:vp_combined}
\end{figure*}

We leverage the viewpoint control probe to conduct fine-grained analysis of how models respond to different reference conditions (\Cref{fig:vp_combined}).

\paragraph{Copy-paste models drift at intermediate viewpoints even with oracle references.}
\Cref{fig:vp_combined}(a,b) shows per-frame identity scores along the camera trajectory.
Under canonical references~(\Cref{fig:vp_combined}a), all models degrade as occluded regions grow toward the target view, which is expected as the viewpoint diverges further from the references; Phantom models drop the least (${\sim}0.05$).
However, the oracle setting~(\Cref{fig:vp_combined}b) reveals a surprising pattern: VACE-14B and SkyReels-A2, the two strongest copy-paste models, exhibit V-shaped curves, dropping at intermediate frames and recovering more near $f_{49}$ where the viewpoint approaches the oracle reference.
This indicates that copy-paste models reproduce the target only when the viewpoint is close but drift at intermediate angles (\Cref{fig:view_inconsistency}), so high oracle scores do not imply multi-view consistency.
SkyReels-A2 makes the distinction concrete: it ranks second conventionally but second-to-last in VC oracle because it recovers near the oracle endpoint while drifting along the intervening trajectory. Thus raw oracle performance is not the appropriate proxy for free-generation reference reuse; the matched canonical-to-oracle gain more directly factors out trajectory-following difficulty.

\paragraph{Not all models benefit from more references.}
\Cref{fig:vp_combined}(c) shows viewpoint control performance as the number of reference images increases from zero (white image) to three.
Phantom-1.3B and Phantom-14B exhibit clear monotonic improvement, gaining ${\sim}0.05$ from 0 to 3 references, with the largest jump at the 0 to 1 transition.
In contrast, several models fail to scale: VACE-14B plateaus after one reference ($+0.03$ from 0 to 1 then flat), Kaleido improves through two references but \emph{drops} at three, and MAGREF decreases when given one reference compared to none before slowly recovering.
Ideally, additional references should provide complementary information and improve performance.
Phantom models show this trend, while the non-monotonic patterns of other models suggest limited multi-reference reasoning: additional views are either ignored or introduce conflicting signals.

\paragraph{Summary.}
Copy-paste models achieve high scores only near the reference viewpoint but drift elsewhere; Phantom models produce more consistent views but do not fully preserve identity at the target view. Neither strategy alone solves the problem, motivating the training-data analysis below.

\subsection{Copying, Generalization, and Training Data Bias}
\label{sec:training}

The copying and generalization behaviors identified in \cref{sec:ref_utilization} are consistent with a key difference in training data construction, discussed below.

\paragraph{Cross-video vs.\ within-clip subject-to-video training.}
While the evaluated models also differ in architecture, conditioning mechanism, and parameter count, the behavioral differences above align most closely with a key distinction in training data construction.
Phantom is the only evaluated model that strictly pairs references and targets across different videos of the same subject~\citep{phantom}: reference images are retrieved from separate clips via CLIP-based matching, guaranteeing that reference and target come from different clips and thus highly likely differ in viewpoint and subject state.
This \emph{cross-video} pairing encourages learning a cross-context mapping: given one view or state of a subject, predict how it appears under a different view or state, which directly mirrors the canonical evaluation setting.
In contrast, models exhibiting the strongest copy-paste behavior (VACE, SkyReels-A2) train with references segmented from the target clip itself (e.g., via SAM2) and then augmented with simple transformations~\citep{vace,skyreelsa2}.
Because the reference and target share the same scene context, these models may learn to reproduce visible surface cues rather than infer held-out content, which is consistent with both their high oracle scores and their sharp degradation under canonical references.
Unconstrained free generation also reveals this behavioral divergence: given a prompt that exactly describes the reference image, VACE-14B tends to reproduce it while Phantom-14B tends to synthesize novel viewpoints (\cref{sec:free_gen_behavior}).

\paragraph{The missing capability: adaptive reference utilization.}
Avoiding copy-paste does not necessarily indicate a better model.
Ideally, a model should adaptively leverage reference content, copying when the references closely match the target and generalizing when they do not.
However, Phantom underperforms copy-paste baselines like VACE-14B under oracle conditions (\cref{tab:main}), suggesting that it may have over-committed to generalization: even when provided with ground-truth frames, it still attempts cross-view prediction rather than direct reproduction.
Both training paradigms thus reflect different inductive biases: within-clip training produces models that are good at copying, while cross-video training produces models that mainly generalize. Neither adapts its strategy to the available reference content.
The core issue exposed by anchored evaluation is not which models copy more, but that none have learned \emph{when} to copy and when to generalize.
Designing training strategies that teach both behaviors, and when to apply each, remains an open challenge.

\section{Conclusion}
\label{sec:conclusion}

This work asks whether subject-driven video generation models truly understand their subjects or merely copy visible cues from references.
Through anchored evaluation, we show that conventional rankings can reverse when models must generalize beyond their references, and that conventional scores strongly correlate with copying gain.
Human preferences and alternative similarity metrics corroborate these findings: anchored rankings align strongly with human judgments, whereas conventional rankings do not.
Fine-grained analysis reveals that no model strikes the right balance: copy-paste models drift at unseen viewpoints, while generalization-oriented models underuse references even when copying is correct.
We trace this dichotomy to training data construction: the core open problem is learning \emph{when} to copy and when to generalize.

\bibliographystyle{plainnat}
\bibliography{main}

\newpage
\appendix

\section{Degradation--Recovery Data Construction}
\label{sec:dr_data_details}

We construct the degradation--recovery evaluation data from Amazon product pages, which provide both product images and demonstration videos for the same item.
The curation proceeds in three stages.

\paragraph{Stage 1: Category-Based Sampling.}
We first sample a broad pool of items spanning approximately 200 product types.
For each item, we collect the associated metadata, product images, and demonstration videos from the product page.

\paragraph{Stage 2: Automated Filtering.}
We perform a coarse pass by extracting keyframes from each video and computing their similarity to the main product images, discarding items with low visual overlap.
For the remaining items, we compute pairwise DINOv2~\citep{oquab2023dinov2} similarity between each video and its corresponding product images, along with Video Training Suitability Scores (VTSS)~\citep{koala36m}, retaining up to 10 candidate videos per item.

\paragraph{Stage 3: Manual Curation.}
We segment the retained videos into clips and manually select those exhibiting meaningful state changes (e.g., a product being opened, assembled, or operated) where the resulting states can be inferred from the available reference images or other video frames.
Each final sample is required to have at least three reference images from which the state changes in the target video can be reasonably anticipated.
This process yields 100 curated reference--video pairs, each associated with a distinct product and video content ID.

\paragraph{Category Coverage.}
The final dataset contains 100 reference--video pairs spanning 67 product categories, covering household items, electronics, furniture, tools, and other everyday products.
The full list is shown below.

\begin{tcolorbox}[colback=gray!5, colframe=gray!50, title=Product Categories, fonttitle=\bfseries, breakable]
\small
3D Printer, ABIS Book, Air Fryer, Backpack, Camcorder, Camera Continuous Light, Camera Film, Camera Support, Cell Phone Holster, Charging Adapter, Cleaning Brush, Coffee Maker, Countertop Oven, Crib, Dehumidifier, Drinking Cup, Drying Rack, Dutch Ovens, Electronic Component Fan, Electronic Learning Toy, Exercise Mat, Faucet, Fire Pit, Fitness Bench, Food Mixer, Food Processor, Food Smoker, Food Storage Container, Furniture, Garment Steamer, Hair Iron, Heat Press, Ice Chest, Ice Maker, Keyboard Mouse Set, Knife Block Set, Lamp, Lawn Mower, Meal Holder, Microphone Stand, Mop Bucket Set, Musical Instruments, Noise Machine, Non Riding Toy Vehicle, Notebook Computer, Paint, Personal Pill Dispenser, Play Workstation, Portable Electronic Device Cover, Portable Electronic Device Stand, Pressure Cooker, Printer, Security Camera, Sewing Machine, Sport Equipment Bag Case, Steam Cleaner, Storage Box, Storage Rack, Template Stencil, Toaster, Toothbrush, Toy Figure Playset, Trash Can, Utility Cart Wagon, Video Game Console, Video Game Controller, Writing Board.
\end{tcolorbox}

\section{Viewpoint Control Data Construction}
\label{sec:vp_data_details}

We construct the viewpoint control evaluation data from OmniObject3D~\citep{omniobject3d}, which provides approximately 6{,}000 scanned 3D objects across 190 categories, each with ground-truth textured meshes, depth maps, and multi-view Blender renders.
The curation proceeds in five stages.

\paragraph{Stage 1: View Disparity Scoring.}
For each object, we select 8 uniformly spaced azimuth views from the Blender renders and compute pairwise DINOv2~\citep{oquab2023dinov2} distances between their embeddings.
Objects with low disparity (e.g., spherically symmetric or uniformly textured) are discarded, as viewpoint control on such objects is trivially easy and cannot differentiate model capabilities.
We also filter objects with insufficient foreground ratio, indicating poor mesh quality or segmentation.

\paragraph{Stage 2: Disparity-Based Ranking.}
We rank the remaining objects by their maximum, mean, and minimum pairwise DINO distances, yielding three complementary orderings that capture different aspects of view-dependent difficulty.
The top candidates from each ranking (216 entries total) are retained for further review.

\paragraph{Stage 3: Visual Sanity Check.}
We render overview grids for the top-ranked objects to verify that high disparity scores reflect genuine geometric or textural variation rather than rendering artifacts (e.g., mesh penetration, shadow baking).

\paragraph{Stage 4: Source--Target View Pair Selection.}
For each retained object, we select a (source, target) view pair satisfying: azimuth difference $\in [35^\circ, 50^\circ]$, elevation difference $\in [15^\circ, 25^\circ]$, and warped-frame hole ratio $< 0.65$.
These constraints ensure the viewpoint change is substantial enough to be non-trivial, while retaining sufficient visible pixels ($\geq$35\%) for the warped frames to provide meaningful geometric guidance rather than degenerating into pure inpainting.

\paragraph{Stage 5: Manual Review.}
A final human review selects 100 objects, balancing viewpoint difficulty with category diversity and filtering out objects with incomplete meshes or poor framing.
For each selected object, we unproject the source image into a point cloud using its ground-truth depth and render this point cloud along a 49-frame interpolated camera trajectory from source to target pose, producing warped frames and per-frame validity masks indicating holes where the source view has no coverage.
Three reference images are sampled from held-out viewpoints not on the warping trajectory.
Selection of the two ``near'' references targets a band $\Delta_\text{tgt} \in [45^\circ, 55^\circ]$ around the target view; among candidates in this band, we select the pair with the largest great-circle distance from each other, placing them on opposite sides of the target.
Selection of the ``far'' reference targets a band $\Delta_\text{src} \in [15^\circ, 30^\circ]$ around the source view.
Together, the three references cover the range of appearances the subject exhibits along the source-to-target trajectory.

\paragraph{Category Coverage.}
The final dataset contains 100 scanned object instances, one from each of 100 source categories, including everyday items, food, toys, figurines, tools, and small appliances.
The descriptive names of the selected objects are listed below; each entry corresponds to one distinct source category.

\begin{tcolorbox}[colback=gray!5, colframe=gray!50, title=Selected Objects (One per Category), fonttitle=\bfseries, breakable]
\small
9v Battery, Analog Desk Clock, Baby High Chair, Baby Monitor Camera, Bird Figurine, Blue-and-White Porcelain Figurine, Broccoli Head, Broccolini, Brussels Sprout, Burrito Wrap, Cactus Figurine, Calzone, Carrot Top Half, Cauliflower Head, Ceramic Mug, Ceramic Teapot, Ceramic Vase, Chocolate Swirl Bun, Coconut Juice Carton, Combination Pliers, Computer Mouse, Corn on the Cob, Cream Tube, Detergent Jug, Die, Digital Kitchen Timer, Donut, Dumbbell, Earplugs, Fabric Strap with Grommets, Fineliner Marker, Fire Extinguisher, Folding Camp Cot, Frying Pan, Green Chili Pepper, Green Mango, Halved Red Onion, Heart-Shaped Candle, Hot Dog, Inflatable Toy Hammer, Instant Noodle Cup, Kids Elephant Watch, Kids Rolling Suitcase, Kids School Satchel, Knit Beanie, Loquat, Lotus Root Section, Microwave Oven, Mini Projector, Miniature Handbag, Miniature Toy House, Moldy Bread Slice, Mooncake, Novelty Party Glasses, Owl-Shaped Toothpick Holder, Peach-Shaped Bowl, Peanut in Shell, Pencil Eraser, Picnic Basket, Pig-Shaped Hair Clip, Pocket Calculator, Pyraminx Puzzle, Rice Cooker, Rubber Dog Ball, Rubber-Coated Work Glove, Sandwich, Scissors, Shampoo Bottle, Single Flower, Small Cabinet, Small Nightstand, Soda Can, Square-Tube Lipstick, Star Anise, Steamed Bun, Stegosaurus Toy, Tape Measure, Tea Sachet, Teddy Bear, Tissue Box, Toothpaste Tube, Toy Bus, Toy Coach Bus, Toy Delivery Scooter, Toy Drum, Toy Helicopter, Toy Police Car, Toy Train, Unripe Pomegranate, Vacuum Flask, Wall Outlet, Wave-Shaped Sculpture, Wiggle Car, Wooden Horse Figurine, Wooden Speaker, Wooden Toy Boat, Wrapped Candy, Yam (Chinese Yam), Zebra Figurine, Zongzi.
\end{tcolorbox}

\section{DR Noise Level Ablation}
\label{sec:sigma_ablation}

\begin{figure}[H]
\centering
\includegraphics[width=0.8\linewidth]{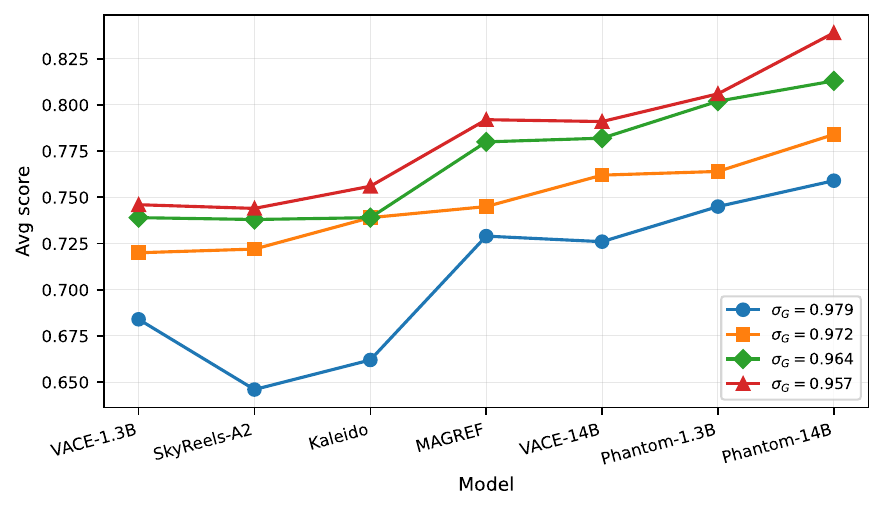}
\caption{DR canonical scores across noise levels. The model ranking is consistent for $\sigma_G \leq 0.972$; at $\sigma_G{=}0.979$ the anchored signal approaches pure noise and rankings become unstable.}
\label{fig:sigma_ablation}
\end{figure}

\Cref{fig:sigma_ablation} shows the DR canonical scores across four noise levels $\sigma_G \in \{0.979, 0.972, 0.964, 0.957\}$, with models sorted by their $\sigma_G{=}0.972$ score (the level used in our main experiments).
The model ranking is highly consistent across $\sigma_G{=}0.972$, $0.964$, and $0.957$: the Pearson/Spearman correlations with $\sigma_G{=}0.972$ are $r{=}0.93$/$\rho{=}0.96$ for $\sigma_G{=}0.964$ and $r{=}0.96$/$\rho{=}0.93$ for $\sigma_G{=}0.957$, indicating that conclusions drawn from our chosen noise level generalize across this range.

The $\sigma_G{=}0.979$ curve shows a notably different pattern, with all models degrading substantially and the relative ranking becoming less stable ($r{=}0.87$, $\rho{=}0.86$).
At this noise level, the anchored latent retains very little structure from the ground truth, so the denoising process receives only weak guidance and the task shifts closer to free generation, diminishing its utility as a test of recovery ability.
This confirms that $\sigma_G$ must be chosen in a range where the anchored signal provides sufficient structure for the task to remain well-defined.

\section{VC Warping Guidance Weight Ablation}
\label{sec:omega_ablation}

\begin{figure}[H]
\centering
\includegraphics[width=0.8\linewidth]{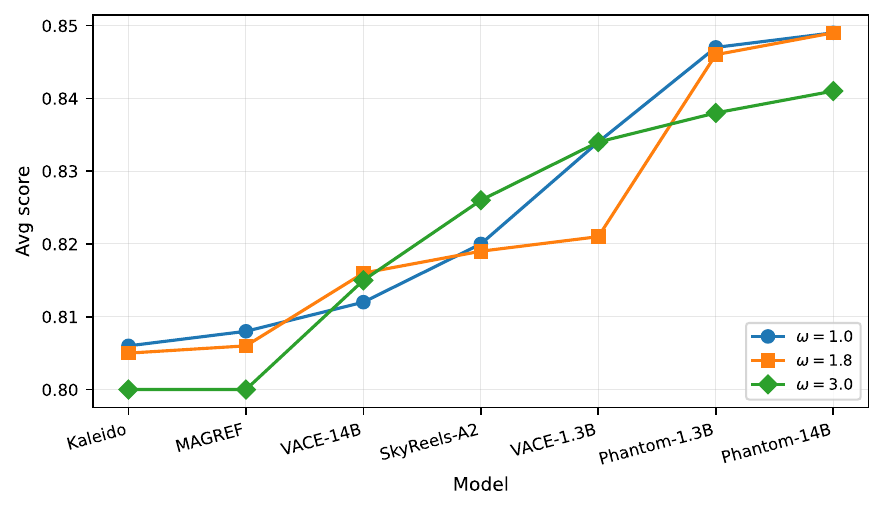}
\caption{VC canonical scores across warping guidance weights. The model ranking is perfectly preserved across all three $\omega$ values ($\rho = 1.00$).}
\label{fig:omega_ablation}
\end{figure}

\Cref{fig:omega_ablation} shows the VC canonical scores across three warping guidance weights $\omega \in \{1.0, 1.8, 3.0\}$, with models sorted by their $\omega{=}1.8$ score (the value used in our main experiments).
The model ranking is perfectly preserved: the Spearman correlation with $\omega{=}1.8$ is $\rho{=}1.00$ for both $\omega{=}1.0$ and $\omega{=}3.0$ (Pearson $r{=}0.96$ and $r{=}0.89$, respectively).
The absolute score differences across $\omega$ values are small (within $\pm$0.015 for most models).
Although different models achieve their best score at different $\omega$ values, the ranking under per-model optimal $\omega$ is identical to that under the fixed $\omega{=}1.8$ used in our main experiments, confirming that the VC probe is robust to this choice.

\section{Human Evaluation Details}
\label{sec:human_details}

We conduct 400 pairwise comparisons in each of five settings---conventional free generation, VC canonical, VC oracle, DR canonical, and DR oracle---for 2,000 comparisons in total. Each trial presents two anonymized videos generated from the same sample and prompt in randomized left--right order with synchronized playback; raters may select either output or a tie. In conventional free generation, raters are shown the subject references and prompt and asked to judge both subject preservation and overall video quality, preferring fewer unnatural copying artifacts when the two outputs otherwise appear comparable. In anchored settings, the held-out ground-truth target is additionally shown, and raters judge which output better preserves the subject under the specified target view or state.

We aggregate pairwise outcomes into model-level Bradley--Terry scores, assigning 0.5 to each model for a tie. \Cref{tab:human_full} reports correlations between automatic scores and human rankings. The conventional score is effectively uncorrelated with human preference, while all four anchored settings show strong Pearson and Spearman correlations.

\begin{table}[H]
\centering
\caption{Human validation across all five evaluation settings. Each row uses 400 pairwise comparisons.}
\label{tab:human_full}
\begin{tabular}{lccc}
\toprule
Setting & Comparisons & Pearson $r$ & Spearman $\rho$ \\
\midrule
Conventional free generation & 400 & $-0.12$ & $0.14$ \\
VC canonical & 400 & $0.77$ & $0.82$ \\
VC oracle & 400 & $0.82$ & $0.87$ \\
DR canonical & 400 & $0.79$ & $0.91$ \\
DR oracle & 400 & $0.81$ & $0.85$ \\
\bottomrule
\end{tabular}
\end{table}

\section{Robustness to Alternative Metrics}
\label{sec:metric_robustness}

We repeat the anchored evaluation with CLIP similarity and, for DR and VC, the identity-focused ID-Sim distance~\citep{idsim2026}. The CLIP ranking agrees closely with the main DINO/GME ranking: Spearman $\rho$ is $0.89/1.00$ for DR canonical/oracle and $0.81/0.85$ for VC canonical/oracle. ID-Sim is a distance, so lower values are better. Its oracle--canonical copying gain remains strongly correlated with conventional performance (Pearson $r{=}0.84$, Spearman $\rho{=}0.96$).

DINO, CLIP, and GME are semantic rather than pixel-level metrics, which is appropriate for stochastic generation where lighting, shadow, scale, and background can change even when a model closely reuses the reference appearance. In the identical-copy limit, reference cosine similarity is maximized; under more realistic processed copying, semantic similarity can remain high even when raw pixels differ. The consistent copying-gain trend across these feature spaces, together with the PSNR analysis below, shows that the finding is neither an artifact of a single embedding space nor solely a semantic-similarity effect.

\begin{table}[H]
\centering
\caption{Per-model anchored results using CLIP similarity (higher is better) and ID-Sim distance (lower is better).}
\label{tab:alternative_metrics}
\scriptsize
\setlength{\tabcolsep}{3.5pt}
\begin{tabular}{lcccccccc}
\toprule
& \multicolumn{4}{c}{CLIP $\uparrow$} & \multicolumn{4}{c}{ID-Sim $\downarrow$} \\
\cmidrule(lr){2-5}\cmidrule(lr){6-9}
Model & DR-can. & DR-ora. & VC-can. & VC-ora. & DR-can. & DR-ora. & VC-can. & VC-ora. \\
\midrule
Phantom-1.3B & .869 & .913 & .864 & .872 & .556 & .319 & .449 & .407 \\
Phantom-14B & .872 & .921 & .872 & .871 & .512 & .287 & .449 & .408 \\
VACE-1.3B & .840 & .881 & .858 & .864 & .734 & .488 & .493 & .378 \\
VACE-14B & .855 & .923 & .861 & .879 & .603 & .283 & .465 & .386 \\
SkyReels-A2-14B & .841 & .910 & .857 & .870 & .737 & .404 & .480 & .419 \\
Kaleido-14B & .856 & .898 & .857 & .864 & .621 & .378 & .539 & .419 \\
MAGREF-14B & .852 & .902 & .856 & .860 & .765 & .482 & .562 & .569 \\
\bottomrule
\end{tabular}
\end{table}

We additionally examine pixel fidelity using the VC PSNR values already reported in \cref{tab:main}. The oracle-minus-canonical PSNR gains are $0.97$, $0.75$, $0.49$, $2.27$, $1.20$, $1.12$, and $0.32$ dB for Phantom-1.3B, Phantom-14B, VACE-1.3B, VACE-14B, SkyReels-A2-14B, Kaleido-14B, and MAGREF-14B, respectively. These gains correlate strongly with conventional performance (Pearson $r{=}0.84$, Spearman $\rho{=}0.89$), reproducing the copying-gain conclusion with a pixel-level metric.

\section{Generalization to Human Faces}
\label{sec:hdtf}

We test whether the anchored ranking extends beyond products and scanned objects using 50 talking-face clips from HDTF~\citep{hdtf2021}, which contain changes in expression, mouth motion, and head pose. For each 49-frame target clip, the canonical setting uses three reference frames outside the target clip with a gap of at least 20 frames, while the oracle setting samples three references from the target clip. We report ArcFace similarity. The canonical and oracle rankings correlate with the corresponding product/object DR rankings at Spearman $\rho{=}0.79$ and $0.89$, respectively.

\begin{table}[H]
\centering
\caption{Anchored evaluation on 50 HDTF talking-face clips using ArcFace similarity.}
\label{tab:hdtf}
\begin{tabular}{lcc}
\toprule
Model & Canonical $\uparrow$ & Oracle $\uparrow$ \\
\midrule
Phantom-14B & .881 & .892 \\
Phantom-1.3B & .878 & .890 \\
VACE-14B & .855 & .904 \\
Kaleido-14B & .848 & .866 \\
SkyReels-A2-14B & .818 & .864 \\
VACE-1.3B & .797 & .822 \\
MAGREF-14B & .762 & .825 \\
\bottomrule
\end{tabular}
\end{table}

\section{Temporal Consistency}
\label{sec:temporal}

The main VC score averages identity similarity at frames 21, 33, and 49. To additionally penalize an intermediate identity drop, we define
\begin{equation}
S_{\mathrm{temp}} = \frac{s_{21}+s_{33}+s_{49}}{3}
- \max\!\left(0,\frac{s_{21}+s_{49}}{2}-s_{33}\right).
\end{equation}

\begin{table}[H]
\centering
\caption{VC oracle temporal-consistency analysis.}
\label{tab:temporal}
\scriptsize
\setlength{\tabcolsep}{5pt}
\begin{tabular}{lcccccc}
\toprule
Model & $s_{21}$ & $s_{33}$ & $s_{49}$ & Mean & Penalty & $S_{\mathrm{temp}}$ \\
\midrule
Phantom-1.3B & .854 & .860 & .866 & .860 & .001 & .859 \\
Phantom-14B & .864 & .858 & .851 & .858 & .000 & .858 \\
VACE-1.3B & .858 & .864 & .849 & .857 & .000 & .857 \\
VACE-14B & .864 & .858 & .873 & .865 & .010 & .855 \\
Kaleido-14B & .860 & .851 & .835 & .849 & .000 & .849 \\
SkyReels-A2-14B & .845 & .823 & .870 & .846 & .034 & .812 \\
MAGREF-14B & .849 & .803 & .787 & .813 & .015 & .798 \\
\bottomrule
\end{tabular}
\end{table}

\section{API-Compatible Closed-Source Comparison}
\label{sec:closed_source}

Exact anchored evaluation requires access to the denoising trajectory, which closed-source APIs do not expose. We therefore construct a matched API-compatible approximation on the same 30 VC canonical samples. Each model receives the warped target video, three subject references, and the text prompt. We sample twice and select the output whose average color is closer to white, following the benchmark's white-background prior. Among evaluated open-source models, only VACE supports this matched input interface. The proxy favors Kling O3 and Seedance 2.0 over VACE, but it cannot control provider-side inference or eliminate artifacts caused by black warped regions and shadows; it should therefore be interpreted as complementary rather than equivalent to exact anchored evaluation.

\begin{figure*}[p]
\centering
\includegraphics[page=1,width=\textwidth]{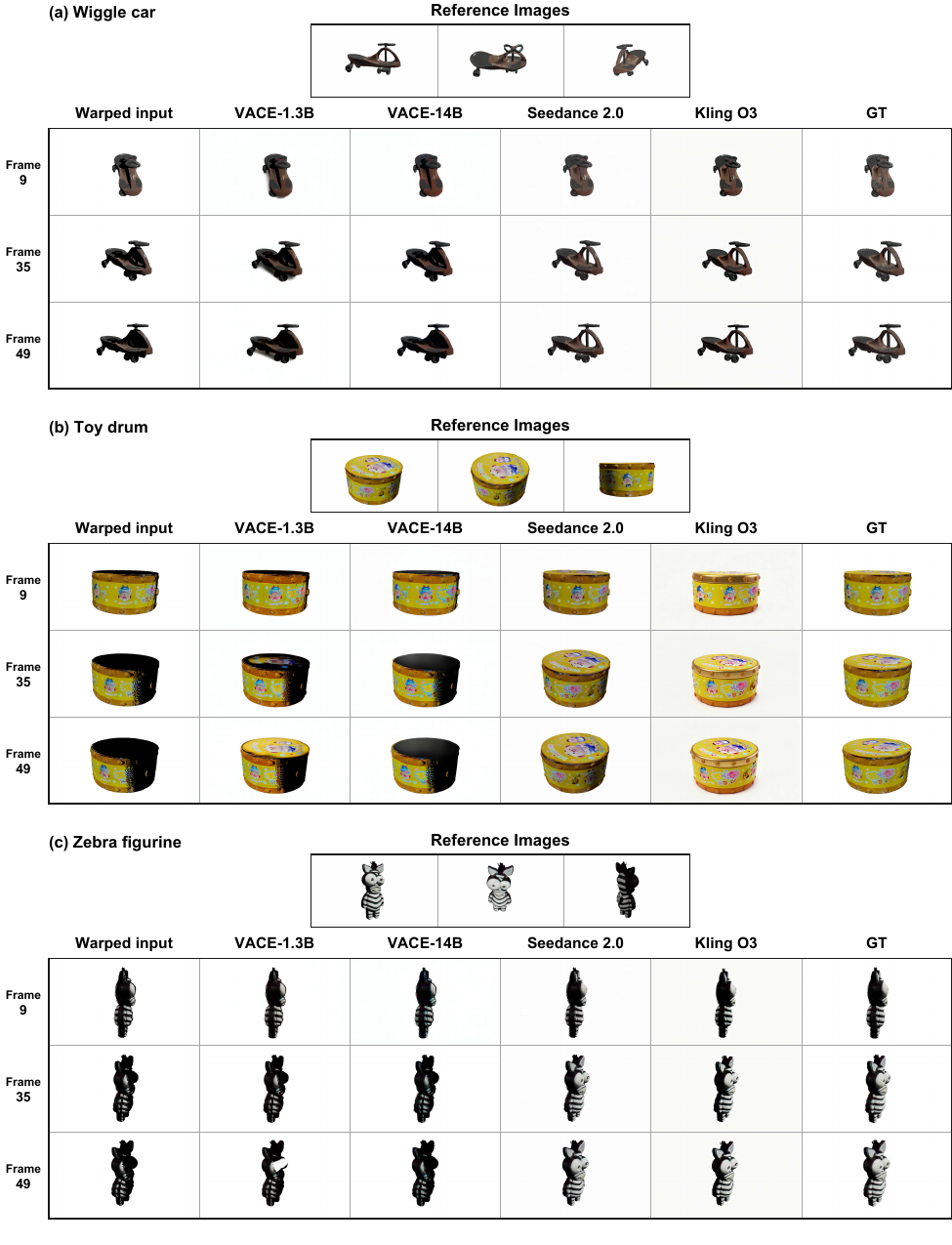}
\caption{Selected success cases under the API-compatible approximation. Seedance 2.0 and Kling O3 are compared with VACE using the same subject references and warped-video conditioning. Rows show frames 9, 35, and 49 after uniformly resampling each clip to 49 frames, aligning normalized clip progress rather than elapsed time. Each displayed run follows the white-background selection rule in Appendix~\ref{sec:closed_source}. These examples illustrate successful appearance completion, not average performance; the API-compatible setting differs from exact anchored evaluation.}
\label{fig:api_success}
\end{figure*}

\begin{figure*}[p]
\centering
\includegraphics[page=2,width=\textwidth]{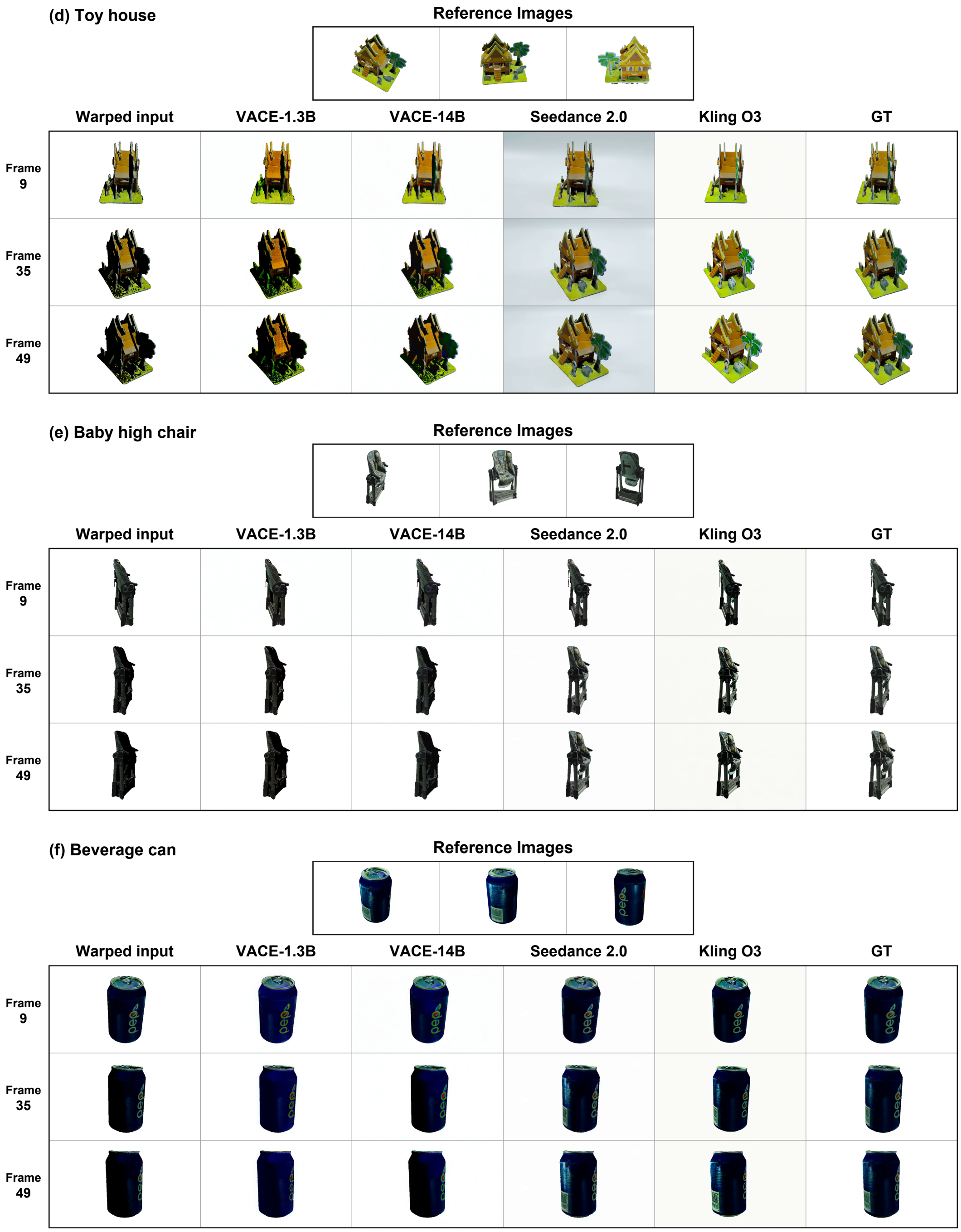}
\caption{Additional selected success cases under the API-compatible approximation, using the same layout, run-selection rule, and normalized frame alignment as Figure~\ref{fig:api_success}. The house structure, high-chair cushion, and can surface illustrate appearance completion by Seedance 2.0 and Kling O3. Original backgrounds and residual deviations are retained; for example, Seedance changes the house background. The task-level prompt template is provided in Appendix~\ref{sec:closed_source}, rather than repeated in the panels.}
\label{fig:api_success_more}
\end{figure*}

\Cref{fig:api_success,fig:api_success_more} show six selected success cases for Seedance 2.0 and Kling O3 alongside the matched VACE baselines. The closed-source outputs recover missing structure and appearance in the car, drum, zebra figurine, toy house, high chair, and beverage can, while the VACE outputs often retain dark regions. These selected cases illustrate successful completion rather than average performance; residual deviations, such as Seedance's background change in the house example, remain visible.

\begin{table}[H]
\centering
\caption{Matched API-compatible comparison on 30 VC canonical samples. Top-1/Top-2 counts are per-sample ranks; rank std. is the standard deviation of ranks.}
\label{tab:closed_source}
\small
\begin{tabular}{lcccccc}
\toprule
Model & DINO & GME & Avg. & Top-1 & Top-2 & Rank std. \\
\midrule
Kling O3 & .831 & .868 & .849 & 12 & 26 & .90 \\
Seedance 2.0 & .815 & .856 & .836 & 14 & 20 & 1.22 \\
VACE-14B & .786 & .811 & .799 & 3 & 10 & 1.02 \\
VACE-1.3B & .769 & .798 & .784 & 1 & 4 & .70 \\
\bottomrule
\end{tabular}
\end{table}

\paragraph{Task-level prompt template.}
The following template expresses the completion request for the API-compatible task. The placeholder \texttt{\{caption\}} is replaced by the sample-specific subject and target-view description. The warped video and three canonical subject-reference images are supplied as separate conditioning inputs; the held-out GT video is used only for evaluation. This is a task-level template, independent of provider-specific request syntax or hidden prompt preprocessing.

\begin{tcolorbox}[colback=gray!5, colframe=gray!50, title=API-Compatible Completion Prompt, fonttitle=\bfseries, fontupper=\small, breakable]
The completed video shows: ``\texttt{\{caption\}}''

\medskip
Complete only the black invisible regions in the input warped video using the three subject reference images. Infer the missing product geometry, surface details, logos, textures, materials, and proportions without copying reference-image patches. Preserve all visible non-black pixels and the pure white background unchanged. Strictly follow the input video's camera trajectory, speed, framing, and novel viewpoints; do not rotate toward the reference-image viewpoints. Maintain seamless spatial completion and temporal consistency throughout the rotation.
\end{tcolorbox}

\section{Comparison with OpenS2V-Nexus}
\label{sec:opens2v_comparison}

We compare our diagnostic with the OpenS2V-Nexus conventional benchmark~\citep{yuan2025opens2v}. Using the DR/VC-averaged scores reported in \cref{tab:main}, NexusScore correlates weakly with canonical and oracle anchored performance (Pearson/Spearman $-0.39/-0.27$ and $0.15/0.25$), but strongly with oracle--canonical copying gain ($0.61/0.85$). It also correlates with our matched conventional free-generation score ($0.69/0.86$). Thus, two conventional protocols agree with one another and both preferentially reward models that benefit from copy-compatible oracle references.

NexusScore is not expected to match the raw oracle score directly. Both NexusScore and our conventional baseline measure reference similarity when the model is free to choose its generation trajectory, whereas an anchored oracle score additionally requires following an imposed video trajectory or viewpoint change. A copy-oriented model may therefore score highly in free generation yet fail along the anchored trajectory, as SkyReels-A2 does at intermediate viewpoints. The closest raw counterpart to NexusScore is consequently our conventional score; the oracle--canonical difference answers a different diagnostic question by holding the anchored trajectory fixed and measuring only how much target-matched references help. This explains why NexusScore agrees strongly with conventional evaluation and with copying gain, but only weakly with either raw anchored score.

\begin{table}[H]
\centering
\caption{OpenS2V-Nexus scores and oracle--canonical copying gains.}
\label{tab:opens2v}
\begin{tabular}{lcc}
\toprule
Model & NexusScore $\uparrow$ & Copying gain $\uparrow$ \\
\midrule
VACE-14B & 44.08 & .104 \\
Kaleido-14B & 43.82 & .066 \\
SkyReels-A2-14B & 43.75 & .086 \\
MAGREF-14B & 43.04 & .062 \\
Phantom-1.3B & 42.48 & .065 \\
VACE-1.3B & 37.91 & .065 \\
Phantom-14B & 37.43 & .053 \\
\bottomrule
\end{tabular}
\end{table}

\section{Free-Generation Behavior: Copying vs.\ Novel-View Synthesis}
\label{sec:free_gen_behavior}

\begin{figure}[t]
\includegraphics[width=1.0\linewidth]{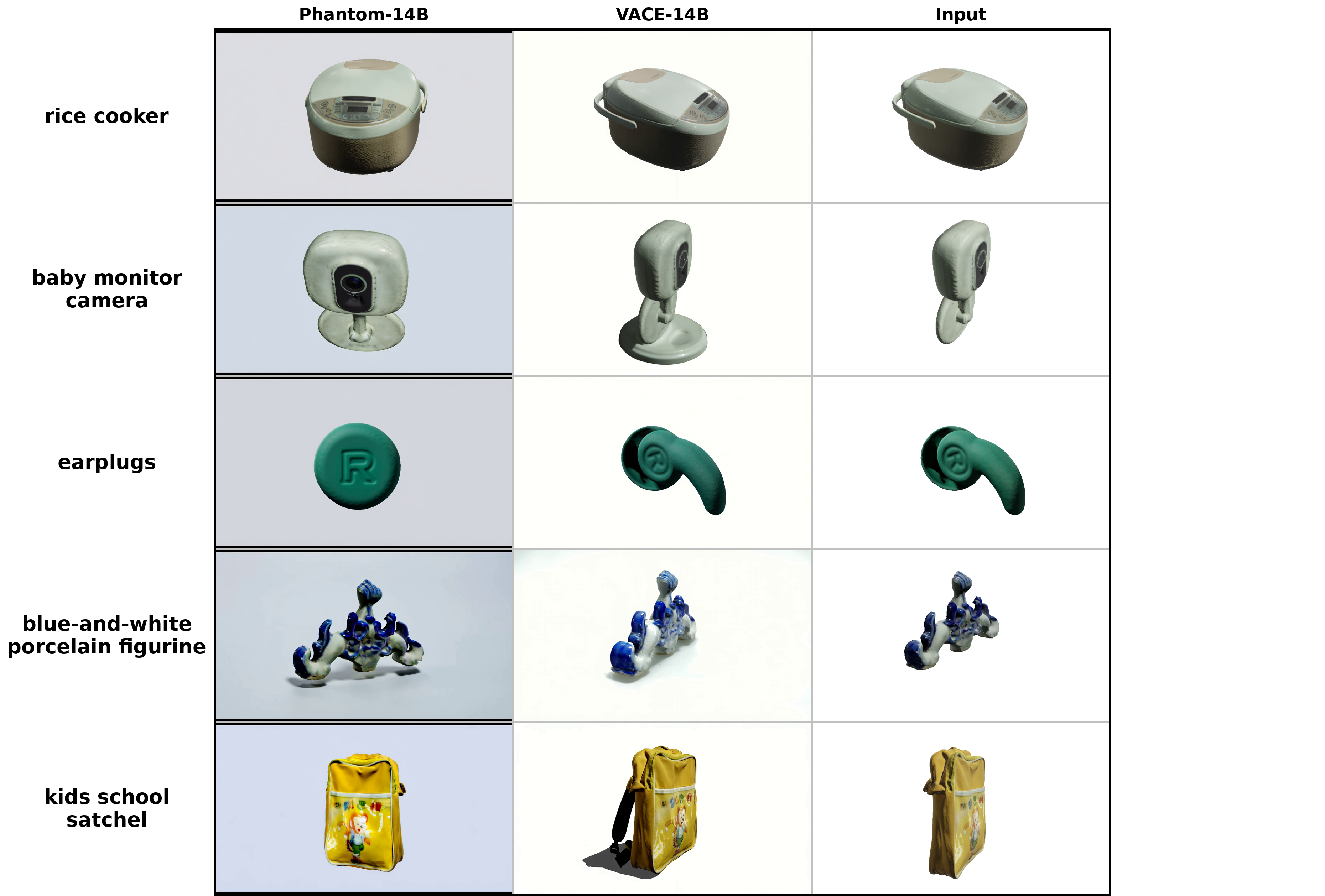}
\caption{Free-generation outputs from Phantom-14B and VACE-14B using the VC target caption (``\texttt{\{category\}} in a pure white background'') as the text prompt and the target view as the sole reference image. VACE-14B tends to reproduce the input with minor refinements (e.g., added shadows or partial object completion) without changing the viewpoint, while Phantom-14B tends to synthesize novel viewpoints even when direct copying would satisfy the prompt.}
\label{fig:free_gen_behavior}
\end{figure}

To further illustrate the training-data-induced behavioral difference discussed in \cref{sec:training}, we run Phantom-14B and VACE-14B in standard free-generation mode using the VC target view as the sole reference and its ground-truth caption (``\texttt{\{category\}} in a pure white background'') as the text prompt (\cref{fig:free_gen_behavior}).
Because the prompt exactly describes the reference image, a model \emph{could} achieve a high similarity score by simply reproducing the input.
VACE-14B tends to replicate the reference viewpoint with only minor refinements such as added drop shadows or partial object completion, without altering the viewpoint.
Phantom-14B, by contrast, tends to synthesize novel viewpoints of the subject, producing different poses and orientations despite having a direct copy-paste path available.
This pattern is consistent with the hypothesis that within-clip training (VACE) biases models toward reproduction, while cross-video training (Phantom) biases models toward novel-view synthesis.

\section{Qualitative Comparison}
\label{sec:qualitative}

\begin{figure*}[p]
\centering
\includegraphics[width=\textwidth]{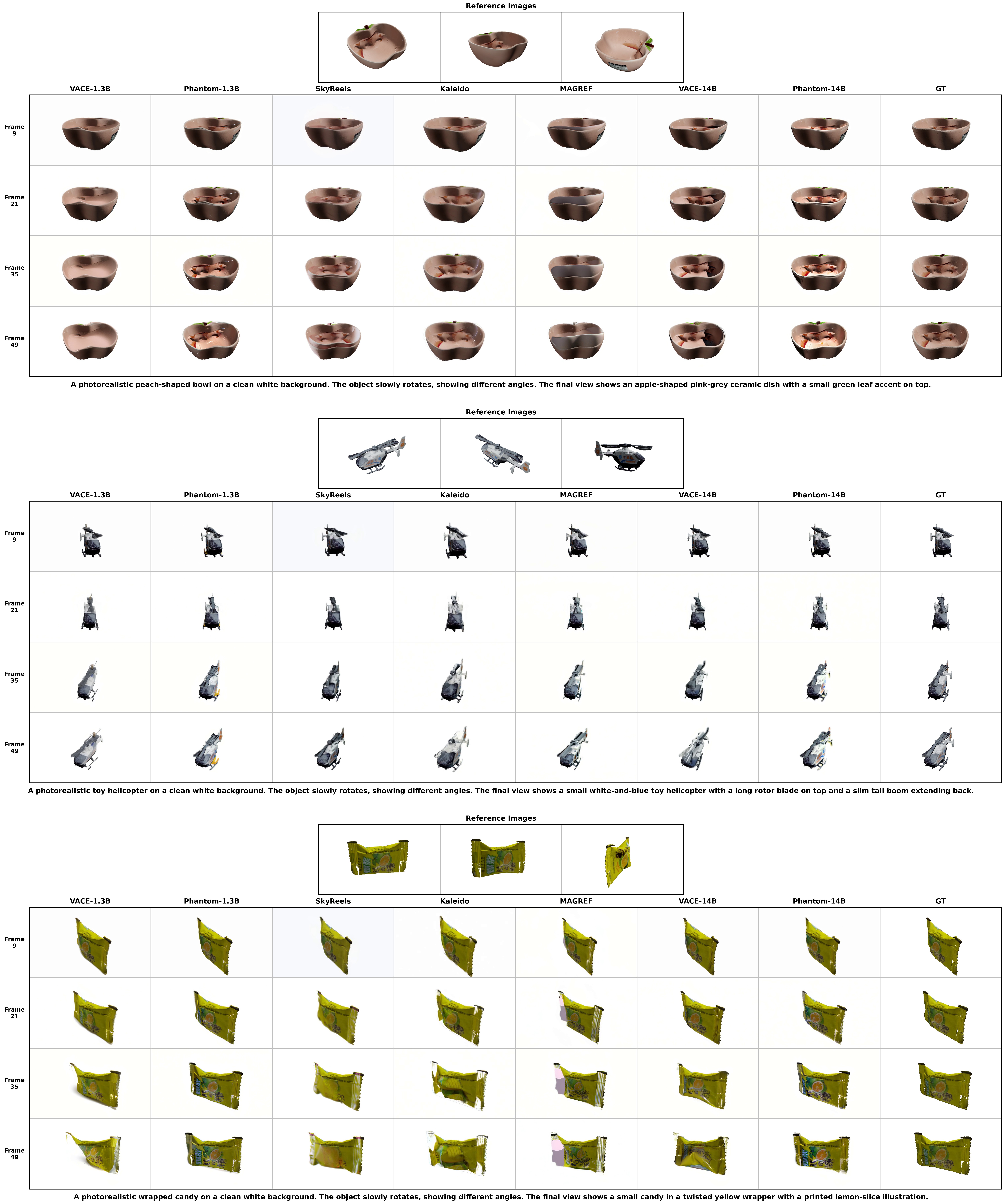}
\caption{Viewpoint control, canonical references. Phantom-14B maintains more consistent subject identity across the camera trajectory.}
\label{fig:vp_canonical}
\end{figure*}

\begin{figure*}[p]
\centering
\includegraphics[width=\textwidth]{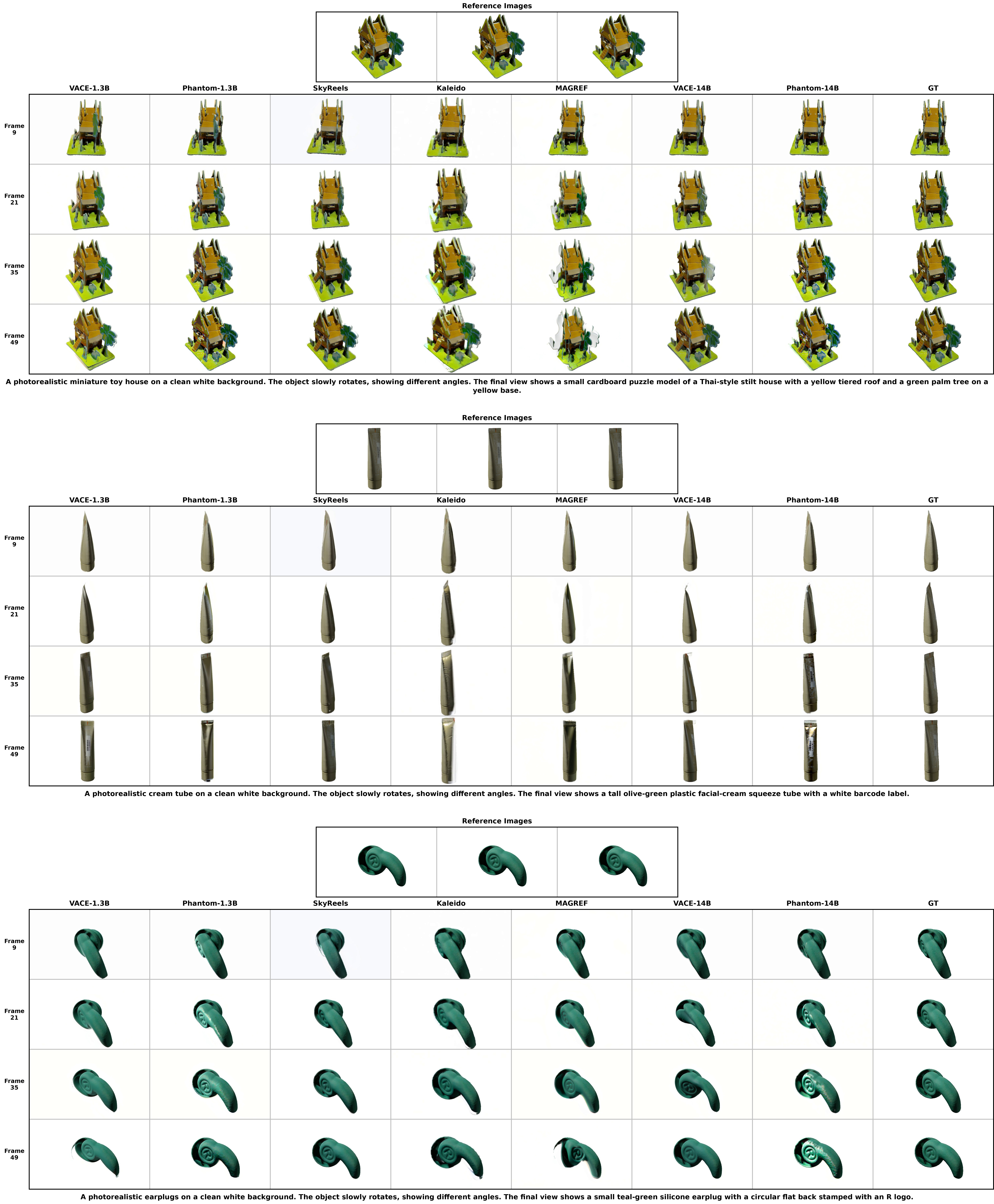}
\caption{Viewpoint control, oracle references. VACE-14B produces sharper results near the target view but may drift at intermediate frames.}
\label{fig:vp_oracle}
\end{figure*}

\begin{figure*}[p]
\centering
\includegraphics[width=\textwidth]{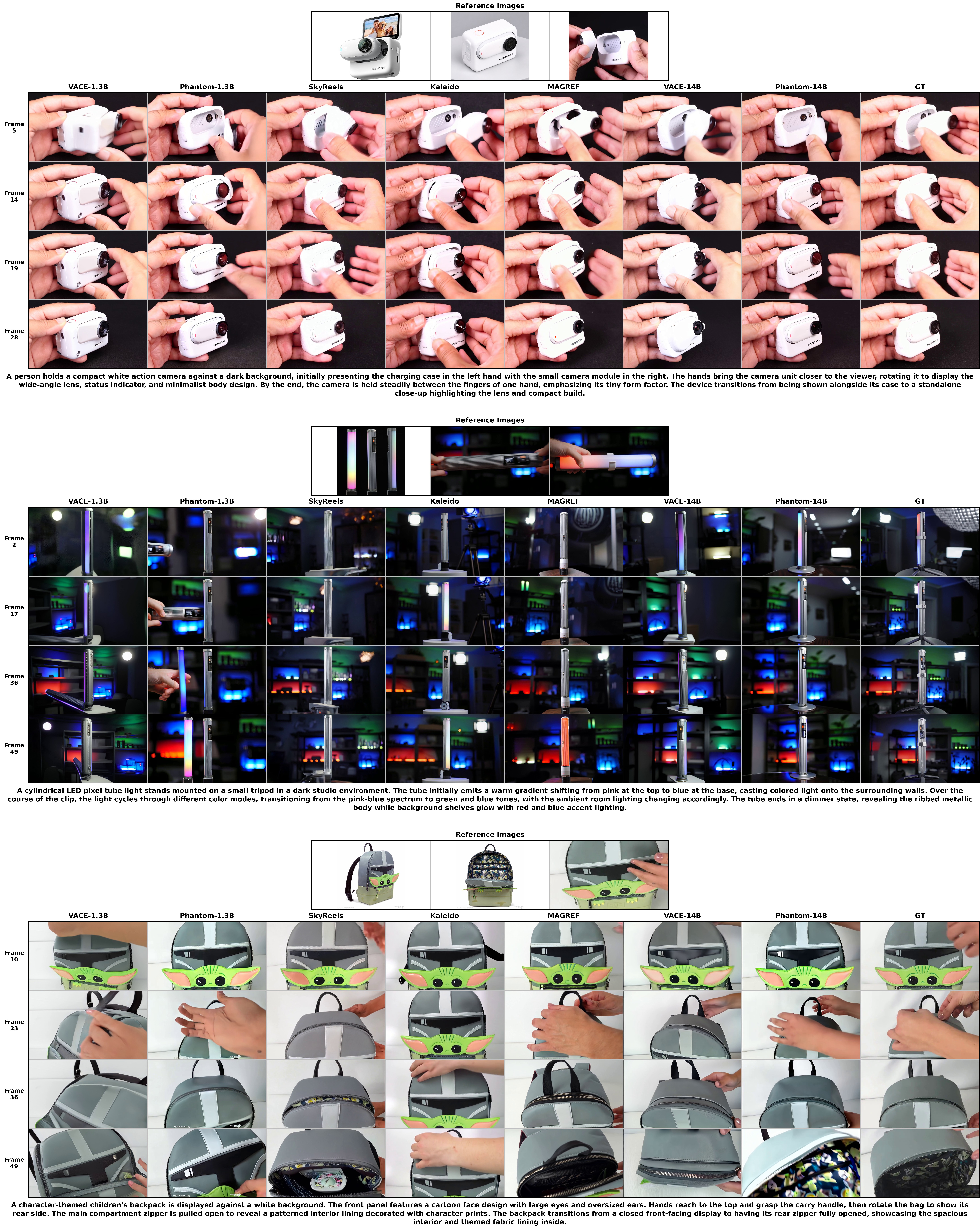}
\caption{Degradation--recovery, canonical references. Phantom-14B better preserves subject appearance when references differ from the target context.}
\label{fig:dr_canonical}
\end{figure*}

\begin{figure*}[p]
\centering
\includegraphics[width=\textwidth]{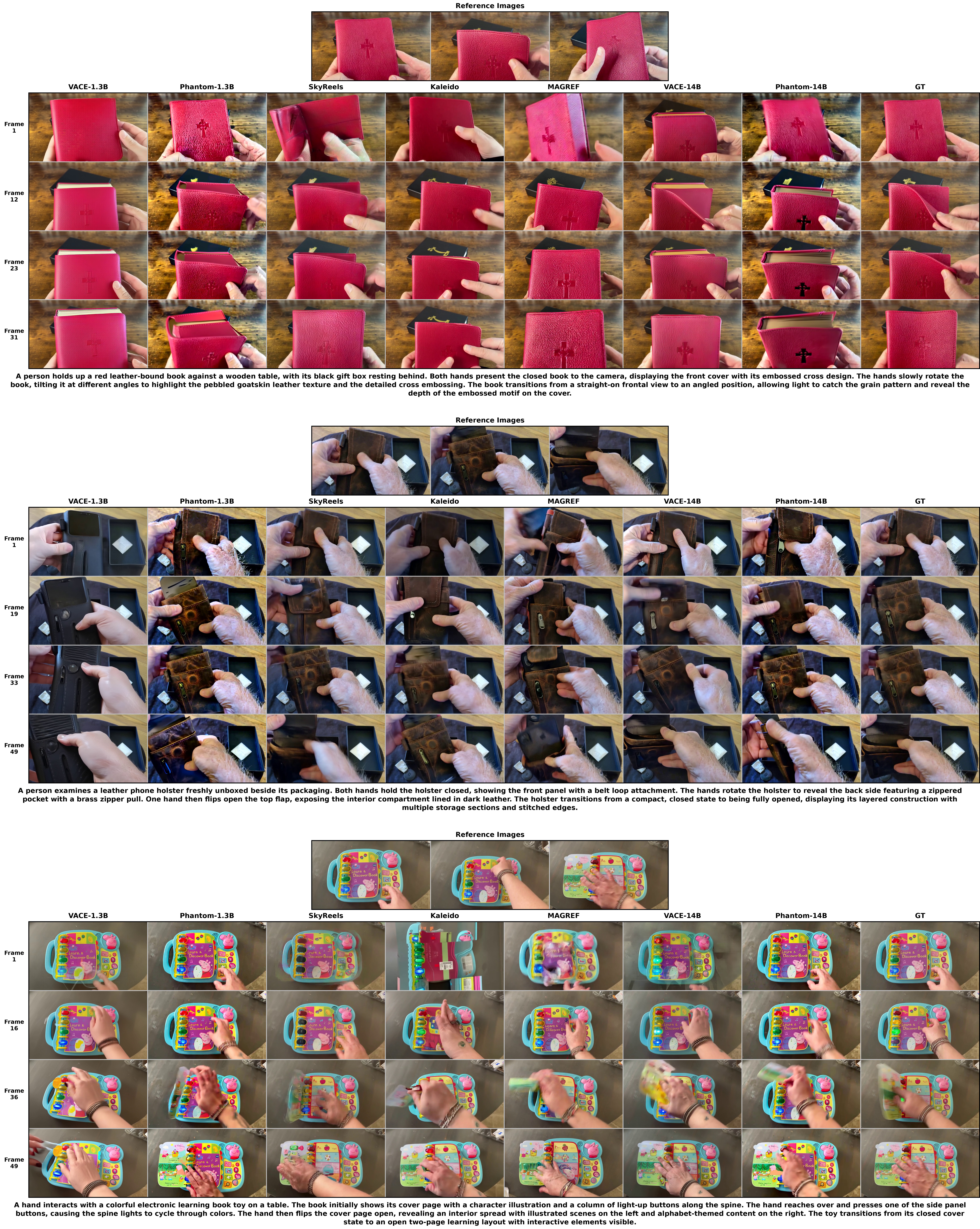}
\caption{Degradation--recovery, oracle references. VACE-14B recovers finer details when ground-truth frames are provided as references.}
\label{fig:dr_oracle}
\end{figure*}

\Cref{fig:vp_canonical,fig:vp_oracle,fig:dr_canonical,fig:dr_oracle} show qualitative comparisons across all seven models under both canonical and oracle reference settings.
The examples visually illustrate the copying-vs.-generalization dichotomy identified in the main paper.

Under canonical references (\cref{fig:vp_canonical,fig:dr_canonical}), where the model must generalize beyond what the references show, Phantom-14B generally produces the most faithful subject identity across both probes, consistent with its top canonical scores in \cref{tab:main}.
Under oracle references (\cref{fig:vp_oracle,fig:dr_oracle}), where direct copying becomes viable, VACE-14B tends to produce sharper and more accurate results, leveraging its strong copy-paste capability.
However, even with oracle references, VACE-14B still exhibits inconsistencies at intermediate viewpoints where the camera is far from the oracle reference (\cref{fig:vp_oracle}), corroborating the V-shaped trajectory in \cref{fig:vp_combined}(b) and confirming that high oracle scores do not imply consistent multi-view understanding.
These patterns reinforce our main finding: no current model adaptively balances copying and generalization across reference conditions.

\section{Limitations}
\label{sec:limitations}

Exact anchored evaluation requires access to the denoising process, so independent evaluation is most faithful for open-source models. Our API-compatible closed-source comparison (Appendix~\ref{sec:closed_source}) provides supporting evidence but cannot exactly reproduce the anchored intervention or control provider-side inference. Closed-source providers can apply the exact framework internally.
Our two probes test viewpoint and degradation recovery but do not cover all aspects of subject understanding (e.g., articulated motion, material properties under novel lighting).
Additionally, our evaluation is currently limited to short videos (49 frames) at 480p resolution; extending anchored evaluation to longer sequences and higher resolutions remains future work.
The central principle---anchoring generation to a held-out target while withholding target identity evidence from subject conditioning---also applies to subject-driven image generation and image-to-image editing. However, those settings require modality-specific anchors and interventions, so validating them is beyond the present video-focused scope.

\section{Broader Impact}
\label{sec:broader_impact}

Our work provides diagnostic tools for understanding subject-driven generation models.
By exposing failure modes that conventional metrics miss, anchored evaluation can guide the development of more reliable and trustworthy generation systems.
We do not foresee direct negative societal impacts from the evaluation framework itself.
However, improved subject-driven generation could raise concerns around deepfakes and unauthorized use of personal likenesses; we hope that better evaluation, by making model capabilities more transparent, can also inform the design of safeguards against such misuse.

\end{document}